\documentclass[11pt]{article}
\usepackage[colorlinks=true,citecolor=blue]{hyperref}
\usepackage{mathptmx, amsmath, amssymb, amsfonts, amsthm, mathptmx, enumerate, color,mathrsfs}
\usepackage{graphicx}
\usepackage{tikz}
\usepackage{multirow}
\usepackage{epstopdf}
\usepackage{multicol}
\usepackage{algorithm}
\usepackage{algorithmicx}
\usepackage{epstopdf}

\newtheorem{theorem}{Theorem}[section]
\newtheorem{lemma}[theorem]{Lemma}

\newtheorem{corollary}[theorem]{Corollary}

\theoremstyle{definition}
\newtheorem{definition}[theorem]{Definition}

\newtheorem{example}[theorem]{Example}

\numberwithin{equation}{section}

\def\oneb{{\bf 1}}

\def\a{{\bf a}}
\def\b{{\bf b}}
\def\c{{\bf c}}

\def\f{{\bf f}}
\def\gbold{{\bf g}}

\def\h{{\bf h}}

\def\x{{\bf x}}

\def\z{{\bf z}}

\def\B{{\mathcal B}}
\def\C{{\mathcal C}}

\def\F{{\mathcal F}}
\def\G{{\mathcal G}}

\def\M{{\mathcal M}}

\def\Nbb{{\mathbb N}}

\def\R{{\mathbb R}}

\def\al{\alpha}
\def\d{\delta}
\def\D{\Delta}
\def\e{\epsilon}
\def\g{\gamma}

\def\l{\lambda}
\def\L{\Lambda}
\def\om{\omega}
\def\OM{\Omega}
\def\r{\rho}
\def\s{\sigma}
\def\SI{\Sigma}
\def\t{\tau}
\def\th{\theta}
\def\Th{\Theta}

\def\balpha{{\boldsymbol \alpha}}
\def\bbeta{{\boldsymbol \beta}}

\def\boldeta{{\boldsymbol \eta}}

\def\bmu{{\boldsymbol \mu}}

\def\bth{{\boldsymbol \theta}}
\def\bphi{{\boldsymbol \phi}}
\def\bpsi{{\boldsymbol \psi}}
\def\bpi{{\boldsymbol \pi}}

\def\bxi{{\boldsymbol \xi}}
\def\bzeta{{\boldsymbol \zeta}}

\def\kai{k \ap \infty}

\def\tai{t \ap \infty}

\def\ap{\rightarrow}

\def\seq{\subseteq}

\def\bi{\{0,1\}}

\def\bz{{\bf 0}}
\def\imp{\; \Longrightarrow \;}

\def\fa{\; \forall}

\def\as{\mbox{ a.s.}}

\def\nm{\Vert}

\renewcommand{\and}{\mbox{$\wedge$}}

\def\bxt{\bxi_{t+1}}
\def\bzt{\bzeta_{t+1}}
\def\bths{\bth^*}

\def\gJ{\nabla J}

\newcommand{\bc}{\begin{center}}
\newcommand{\ec}{\end{center}}
\newcommand{\be}{\begin{equation}}
\newcommand{\ee}{\end{equation}}
\newcommand{\bd}{\begin{displaymath}}
\newcommand{\ed}{\end{displaymath}}
\newcommand{\ba}{\begin{array}}
\newcommand{\ea}{\end{array}}
\newcommand{\ben}{\begin{enumerate}}
\newcommand{\een}{\end{enumerate}}
\newcommand{\bit}{\begin{itemize}}
\newcommand{\eit}{\end{itemize}}
\newcommand{\beq}{\begin{eqnarray}}
\newcommand{\eeq}{\end{eqnarray}}
\newcommand{\btab}{\begin{tabular}}
\newcommand{\etab}{\end{tabular}}
\newcommand{\bfig}{\begin{figure}}
\newcommand{\efig}{\end{figure}}
\newcommand{\btp}{\begin{tikzpicture}}
\newcommand{\etp}{\end{tikzpicture}}

\newcommand{\nmeu}[1]{ \nm #1 \nm_2 }
\newcommand{\nmeusq}[1]{ \nm #1 \nm_2^2 }
\newcommand{\nmi}[1]{ \nm #1 \nm_\infty}

\def\nmsl1{\nm_{{\rm SL1}}}

\def\gJ{\nabla J}
\def\gJt{\gJ(\bth_t)}

\def\nuit{\nu^{-1}(\t)}

\begin{document}

\title
{Recent Advances in Stochastic Approximation with\\
Applications to Optimization and Fixed Point Problems}
\author{
Rajeeva L.\ Karandikar and  M.\ Vidyasagar
\thanks{RLK is with the Chennai Mathematical Institute; rlk@cmi.ac.in.
MV is with the Indian Institute of Technology Hyderabad;
m.vidyasagar@iith.ac.in.
The research of MV was supported by the Science and Engineering Research
Board, India.
}
}

\maketitle

\begin{abstract}
We begin by briefly surveying some results
on the convergence of the Stochastic Gradient Descent (SGD) Method, proved in a
companion paper by the present authors.
These results are based on viewing SGD as a version of Stochastic
Approximation (SA).
Ever since its introduction in the classic paper of Robbins and Monro
in 1951, SA has become a standard tool
for finding a solution of an equation of the form $\f(\bth) = \bz$,
when only noisy measurements of $\f(\cdot)$ are available.
In most situations, \textit{every component} of the putative solution
$\bth_t$ is updated at each step $t$.
In some applications in Reinforcement Learning (RL),
\textit{only one component} of $\bth_t$ is updated at each $t$.
This is known as \textbf{asynchronous} SA.
In this paper, we study \textbf{Block
Asynchronous SA (BASA)}, in which, at each step $t$, \textit{some
but not necessarily all} components of $\bth_t$ are updated.
The theory presented here embraces both conventional (synchronous) SA
as well as asynchronous SA, and all in-between possibilities.
We provide sufficient conditions for the convergence of BASA, and also
prove bounds on the \textit{rate} of convergence of $\bth_t$
to the solution.
For the case of conventional SGD, these results reduce to those
proved in our companion paper.
Then we apply these results to the problem of finding a fixed point
of a map with only noisy measurements.
This problem arises frequently in RL.
We prove sufficient conditions for convergence as well as estimates
for the rate of convergence.
\end{abstract}






\section{Introduction}\label{sec:Intro}

\subsection{Background}\label{ssec:11}

Ever since its introduction in the classic paper of Robbins and Monro
\cite{Robbins-Monro51},
Stochastic Approximation (SA) has become a standard tool in many
problems in applied mathematics.
It is worth noting that the phrase ``Stochastic Approximation''
was coined in \cite{Robbins-Monro51}.
As stated in \cite{Robbins-Monro51}, the original problem formulation in SA
was to find a solution to an equation of the form\footnote{$^1$For the
convenience of the reader, all results cited from the literature are
stated in the notation used in the present paper, which may differ from
the original paper.}
\bd
f(\th) = c ,
\ed
where $f: \R \ap \R$, $c$ is a specified constant, and one has access
only to noisy measurements of the function.
Obviously, one can redefine $f$ and assume that $c = 0$, without 
loss of generality.
Almost at once, the approach was extended to finding a stationary
point of a $\C^1$-function $J(\cdot) : \R \ap \R$ in \cite{Kief-Wolf-AOMS52},
and to the case where $J(\cdot) : \R^d \ap \R$ in \cite{Blum54}.
Other early contributions are \cite{Dvoretzky56,Derman-Sacks-AOMS59}.
%
In the early papers, SA was analyzed under extremely stringent
assumptions on the function, and on the measurement error.
With the passage of time, subsequent researchers have substantially
relaxed the assumptions.

Over the years, SA has become a standard tool for analyzing the behavior of
stochastic algorithms in a variety of areas, out of which two topics
are the focus
in the present paper, namely: optimization, and finding
a fixed point of a contractive map, which arises frequently in
Reinforcement Learning (RL).
The aim of the present paper is two-fold:
First, we survey some known results in the theory of SA, including
some results due to the present authors.
Second, we present some new results on so-called Block Asynchronous SA,
or BASA.

\subsection{Problem Formulation}\label{ssec:12}

Suppose $\f : \R^d \ap \R^d$ is some function.
It is desired to find a solution to the equation $\f(\bths) = \bz$,
when only noisy measurements of $\f(\cdot)$ are available.
An iterative approach is adopted to solve this equation.
Let $t$ denote the iteration count, and choose the initial guess $\bth_0$
either in a deterministic or a random fashion.
At time (or step) $t+1$, the available measurement is $\f(\bth_t) + \bxt$,
where $\bxt$ is variously referred to as the measurement error
or the ``noise.''
Both phrases are used interchangeably in this paper.
The current guess $\bth_t$ is updated via the formula
\be\label{eq:121}
\bth_{t+1} = \bth_t + \balpha_t \circ [ \f(\bth_t) + \bxt ],
\ee
where $\balpha_t \in (0,\infty)^d$ is called the \textbf{step size vector},
and $\circ$ denotes the Hadamard product.\footnote{$^2$Recall
that if $\a,\b$ are vectors of equal dimension, then their \textbf{Hadamard
product} $\c = \a \circ \b$ is defined by $c_i := a_i b_i$ for all $i$.}
If $\gbold : \R^d \ap \R^d$ is a map and it is desired to find a fixed
point of $\gbold(\cdot)$, when we can define $\f(\bth) = \gbold(\bth) - \bth$.
This causes \eqref{eq:121} to become
\be\label{eq:122}
\bth_{t+1} = ( \oneb_d - \balpha_t) \circ \bth_t +
\balpha_t \circ [ \gbold(\bth_t) + \bxt ] ,
\ee
where $\oneb_d$ denotes the column vector of $d$ ones.
In this case, it is customary to restrict $\balpha_t$ to belong to
$(0,1)^d$ instead of $(0,\infty)^d$.
Then each component of $\bth_{t+1}$ is a convex combination of the
corresponding components of $\bth_t$ and the noisy measurement of
$\gbold(\bth_t)$.
If $J : \R^d \ap \R$ is a $C^1$-function, and it is desired to find
a stationary point of it, then we can define $\f(\bth) = - \gJ(\bth)$,
in which case \eqref{eq:121} becomes
\be\label{eq:123}
\bth_{t+1} = \bth_t + \balpha_t \circ [ - \gJ(\bth_t) + \bxt ] .
\ee
The choice $\f(\bth) = -\gJ(\bth)$ instead of $\gJ(\bth)$ is used
when the objective is to \textit{minimize} $J(\cdot)$, and $J(\cdot)$
is convex, at least in a neighborhood of the minimum.
If the objective is to maximize $J(\cdot)$, then one would choose
$\f(\bth) = \gJ(\bth)$.

What is described above is the ``core'' problem formulation.
Several variations are possible, depending on the objective of the analysis,
the nature of the of the step size vector, and the nature of
the error vector $\bxt$.
Some of the most widely studied variations are described next.

\textbf{Objectives of the Analysis:}
Historically,
the majority of the literature is devoted to showing that
the iterations converge \textit{in expectation} to a solution of  
the equation $\f(\bth) = \bz$ (or its modification for fixed point
and stationarity problems).
This is the objective in \cite{Kushner-JMAA77} and other subsequent papers.
In recent times, the emphasis has shifted towards proving that the
iterations converge \textit{almost surely} to the desired limit.
Since any stochastic algorithm such as \eqref{eq:123} generates
\textit{a single sample path}, it is very useful to know that almost
every run of the algorithm leads to the desired outcome.

Another possibility is \textbf{convergence in probability}.
Suppose $\bth_t \ap \bths$ in probability, and define
\be\label{eq:124}
q(t,\e) := \Pr \{ \nmeu{\bth_t - \bths} > \e \} .
\ee
The objective is to derive suitable conditions under which,
$q(t,\e) \ap 0$ as $\tai$ for each $\e > 0$, and if possible,
to derive explicit upper bounds for $q(t,\e)$.
Some authors refer to such bounds as ``high probability bounds.''
The advantage of bounds on $q(t,\e)$ is that they are applicable
for \textit{all} $t$ (or at least, for all \textit{sufficiently large} $t$),
and not just when $\tai$.
For this reason, some authors refer to the derivation of such bounds as
\textbf{finite-time SA}.
Some contributions in this direction are
\cite{MJW-arxiv19a,Srikant-Ying19,BRS-CoLT18,CMSS-arxiv2102,Qu-Wierman-CoLT20}.
We do not discuss FTSA in the paper.
The interested reader is referred to the above-cited papers and the
references therein.

\textbf{Step Size Sequences:}
Next we discuss various options for the step size vector $\balpha_t$,
which is allowed to be random.
In all cases, it is assumed that there is a \textit{scalar deterministic}
sequence $\{ \beta_t \}$ taking values in $(0,\infty)$,
or in $(0,1)^d$ in the case of \eqref{eq:122}.
We will discuss three commonly used variants of SA, namely: synchronous
(also called fully synchronous), asynchronous, and block asynchronous.
In \textbf{synchronous SA}, one chooses $\balpha_t = \beta_t  \oneb_d$.
Thus, in \eqref{eq:121}, \textit{the same} step size $\beta_t$ is
applied to \textit{every} component of $\bth_t$.
In block asynchronous SA (or BASA), there are $d$ different $\bi$-valued
stochastic processes, denoted by $\kappa_t^i, i \in [d]$, called the
``update'' processes.
Then the $i$-th component of $\bth_t$ is updated only if $\kappa_t^i = 1$.
To put it another way, define the ``update set'' as
\bd
S_t := \{ i \in [d] : \kappa_t^i = 1 \} .
\ed
Then $\al_t^i = 0$ if $i \not\in S_t$.
However, this raises the question as to what $\al_t^i$ is for $i \in S_t$.
Two options are suggested in the literature, known as the ``global''
clock and the ``local'' clock respectively.
This distinction was first suggested in \cite{Borkar98}.
If a global clock is used, then $\al_t^i = \beta_t$.
To define the step size when a local clock is used, first define
\be\label{eq:125}
\nu_t^i := \sum_{\t = 0}^t \kappa_t^i .
\ee
Thus $\nu_t^i$ counts the number of times that $\th_t^i$ is
updated, and is referred to as the ``counter'' process.
Then the step size is defined as
\be\label{eq:126}
\al_t^i := \beta_{\nu_t^i} .
\ee
The distinction between global and local clocks can be briefly summarized
as follows:
When a global clock is used, every component of $\bth_t$ that gets updated
has exactly the same step size, namely $\beta_t$, while the other
components have a step size of zero.
When a local clock is used, among the components of $\bth_t$ that get
updated at time $t$, different components may have different step sizes.
An important variant of BASA is asynchronous SA (ASA).
This phrase was apparently first used in \cite{Tsi-ML94}, in the
context of proving the convergence of the $Q$-learning algorithm
from Reinforcement Learning (RL).
In ASA, \textit{exactly one} component of $\bth_t$ is updated at
each $t$.
This can be represented as follows:
Let $\{ N_t \}$ be an integer-valued stochastic process taking values in $[d]$.
Then, at time $t$, the update set $S_t$ is the singleton $\{ N_t \}$.
The counter process $\nu_t^i$ is now defined via
\bd
\nu_t^i = \sum_{\t = 0}^t I_{ \{ N_\t = i \} } ,
\ed
where $I$ denotes the indicator process.
The step size can either be $\beta_t$ if a global clock is used,
or $\beta_{\nu_t^i}$ if a local clock is used.
In \cite{Borkar98}, the author analyzes the
convergence of ASA with both global as well as local clocks.
In the $Q$-learning algorithm introduced in \cite{Watkins-Dayan92},
the update is asynchronous (one component at a time) and a global
clock is used.
In \cite{Tsi-ML94}, where the phrase ASA was first introduced,
the convergence of ASA is proved under some assumptions which include
$Q$-learning as a special case.
Accordingly, the author uses a global clock in the formulation of ASA.
In \cite{Dar-Mansour-JMLR03}, the authors use a local clock
to study the rate of convergence of $Q$-learning.

\textbf{Error Vector:}
Next we discuss the assumptions made on the error vector $\bxt$.
To state the various assumptions precisely, let $\bth_0^t$ denote
$( \bth_0 , \cdots , \bth_t )$, and define $\balpha_0^t$ and
$\bxi_1^t$ analogously; note that there is no $\bxi_0$.
Let $\F_t$ denote the $\s$-algebra generated by $\bth_0, \balpha_0^t,
\bxi_1^t$, and observe that $\F_t \seq \F_{t+1}$.
Thus $\{ \F_t \}_{t \geq 0}$ is a filtration;
Now \eqref{eq:121} makes it clear that $\bth_t$ is measurable
with respect to $\F_t$, denoted by $\bth_t \in \M(\F_t)$.
Given an $\R^d$-valued random variable $X$, let $E_t(X)$ denote $E(X|\F_t)$,
the conditional expectation of $X$ with respect to $\F_t$,
and let $CV_t(X)$ denote the conditional variance of $X$, defined as
\bd
CV_t(X) = E_t( \nmeusq{X - E_t(X)} ) = E_t(X^2) - [E_t(X)]^2 .
\ed
An important ingredient in SA theory is the set of assumptions imposed on
the two entities $E_t(\bxt)$ and $CV_t(\bxt)$.
We begin with $E_t(\bxt)$,
The simplest assumptions are that
\be\label{eq:127}
E_t(\bxt) = \bz , \fa t ,
\ee
and that there exists a constant $M$ such that
\be\label{eq:128}
CV_t(\bxt) \leq M , \fa t .
\ee
where the equality and the bound hold almost surely.
To avoid tedious repetition, the phrase ``almost surely'' is omitted
hereafter, unless it is desirable to state it explicitly.
Equation \eqref{eq:127} implies that $\{ \bxi_t \}$ is a martingale
difference sequence with respect to the filtration $\{ \F_t \}$.
Equation \eqref{eq:127} further means that $\f(\bth_t) + \bxt$
provides an \textit{unbiased} measurement of $\f(\bth_t)$.
In \eqref{eq:129}, the bound on $CV(\bxt)$ is not just uniform over $t$,
\textit{but also uniform over $\bth_t$}.
Over time, the assumptions on both $E_t(\bxt)$ and $CV_t(\bxt)$
have been relaxed by successive authors.
The most general set of conditions to date are found in
\cite{MV-RLK-SGD-arxiv23},\footnote{$^3$This
paper is currently under final review by \textit{Journal of
Optimization Theory and Applications.}}
and are as follows:
There exist sequences of constants $\mu_t$ and $M_t$ such that
\be\label{eq:129}
\nmeu{ E_t(\bxt ) } \leq \mu_t ( 1 + \nmeu{\bth_t} ) , \fa t .
\ee
\be\label{eq:1210}
CV_t(\bxt) \leq M_t ( 1 + \nmeusq{\bth_t} ) , \fa t .
\ee
In \cite{MV-RLK-SGD-arxiv23}, the following are established:
\ben
\item Suppose
\bd
\sum_{t=0}^\infty \al_t^2 < \infty , 
\sum_{t=0}^\infty \al_t \mu_t  < \infty ,
\sum_{t=0}^\infty \al_t^2 M_t^2 < \infty .
\ed
Then the iterations $\{ \bth_t \}$ are bounded almost surely.
\item If in addition
\bd
\sum_{t=0}^\infty \al_t = \infty ,
\ed
then $\bth_t$ converges almost surely to the unique solution of \eqref{eq:121}.
\een
Thus, by suitably tuning the step size sequence, bounds of
the form \eqref{eq:129} and \eqref{eq:1210} can be accommodated.
The literature review in \cite[Section 1.1]{MV-RLK-SGD-arxiv23}
contains details of the various intermediate stages between
\eqref{eq:127}--\eqref{eq:128} and \eqref{eq:129}--\eqref{eq:1210},
and the relevant publications.
A  condensed version of it is reproduced in Section \ref{ssec:21}.
The reader is also directed to \cite{Lai03} for a partial survey
that is up to date until its date of publication, 2003.

\textbf{Methods of Analysis:}
There are two broad approaches to the analysis of SA,
which might be called the ODE approach and the martingale approach.
In the ODE approach, it is shown that, as the step sizes $\al_t \ap 0$,
the stochastic sample paths of \eqref{eq:121} ``converge'' to the
(deterministic) solution trajectories of the associated ODE
$\dot{\bth} = \f(\bth)$.
This approach is introduced in \cite{Kushner-JMAA77,Ljung-TAC77b,Der-Fradkov74}.
Book-length treatments of the ODE approach can be found in
\cite{Kushner-Clark12,Kushner-Yin03,BMP92,Borkar22}.
The Kushner-Clark condition \cite{Kushner-Clark12}
is not a directly verifiable condition,
but one needs to fall back on martingale or similar
assumptions (such as ‘mixingale’) on noise to verify it.
The martingale method was pioneered in \cite{Gladyshev65},
and independently discovered and enhanced in \cite{Robb-Sieg71}.
In this approach, the stochastic process $\{ \bth_t \}$ is directly analyzed
without recourse to any ODE.
Conclusions about the behavior of this stochastic process are drawn
using the theory of supermartingales.
The two methods complement each other.
A typical theorem based on the ODE approach states that
\textit{if} the iterations remain bounded almost surely,
then convergence takes place.
Often the boundedness (also called ``stability'') can be established
using other methods.
Also, the ODE approach can address the situation where the equation
has multiple solutions.
In contrast, in the martingale approach, both the boundedness and the
convergence of the iterations can be established simultaneously.
An important paper in the ODE approach is \cite{Borkar-Meyn00},
in which the boundedness of the iterations is a conclusion and not
a hypothesis.

\subsection{Contributions of the Paper}\label{ssec:13}

After  the survey of the Stochastic Gradient method,
the emphasis in the paper is on the finding the solution of a fixed-point
equation of the following form:
Suppose $\h$ maps the sequence space $(\R^d)^\Nbb$ into itself.
The objective is to find a fixed point $\x^* \in (\R^d)^\Nbb$ such that
\be\label{eq:1211}
\h_t(\x^*) = \x_t^* , \fa t \geq 0 .
\ee
This part of the paper consists of an analysis of Block (or Batch)
Asynchronous SA, or BASA, for finding a solution to \eqref{eq:1211}.
Suppose $\h(\cdot)$ is a memoryless contraction, in the sense that
\bd
\h_t(\x) = \gbold(\x_t) 
\ed
for some map $\gbold : \R^d \ap \R^d$ which is a contraction in the 
$\ell_\infty$-norm.
Then the formulation reduces to \eqref{eq:122}.
But we also the more general case where $\h$ has memory, delays, etc.
Towards this end, we begin by analyzing the
convergence of ``intermittently updated'' processes of the form
\bd
w_{t+1} = (1 - \al_t \kappa_t) w_t + \al_t \kappa_t \xi_{t+1} ,
\ed
where $\{ w_t \}$ is an $\R$-valued stochastic process,
$\{ \xi_t \}$ is the measurement error,
$\{ \al_t \}$ is a $(0,1)$-valued ``step size'' process,
and $\{ \kappa_t \}$ is a $\bi$-valued ``update'' process.
For this formulation, we derive sufficient conditions for convergence,
as well as bounds on the \textit{rate} of convergence.
We study both the use of both a local clock as well as a global clock,
a distinction first introduced in \cite{Borkar98}.
This formulation is a precursor to the full BASA formulation of
\eqref{eq:122}, where again we derive both sufficient conditions for
convergence, and bounds on the \textit{rate} of convergence.

\subsection{Scope and Organization of the Paper}\label{ssec:14}

This paper contains a survey of some results due to the present authors,
and some new results.
In Section \ref{sec:SSA}, various results from \cite{MV-RLK-SGD-arxiv23}
are stated without proof; these results pertain to the convergence of
the synchronous SA algorithm, when the error signal $\bxt$ satisfies
the bounds \eqref{eq:129} and \eqref{eq:1210}.
These are the most general assumptions to date.
In Section \ref{sec:Opt}, we survey some applications of these
convergence results to the stochastic gradient method.
The results in \cite{MV-RLK-SGD-arxiv23} make the least
restrictive assumptions on the measurement error.
These two sections comprise the survey part of the paper.

In Section \ref{sec:BASA}, we commence presenting some new results.
Specifically, we study Block (or Batch) Asynchronous SA, denoted by BASA,
as described in \eqref{eq:122}.
The focus is on finding a fixed point of a map $\gbold: \R^d \ap \R^d$
which is a contraction in the $\ell_\infty$-norm, or a scaled version thereof.
While this problem arises in Reinforcement Learning in several situations,
finding fixed points is a pervasive application of stochastic approximation.
The novelties here are that (i) we permit a completely general model
for choosing the coordinates of $\bth_t$ to be updated at time $t$, and
(ii) we also derive bounds on the \textit{rate} of convergence.

\section{Synchronous Stochastic Approximation}\label{sec:SSA}

\subsection{Historical Review}\label{ssec:21}

We begin with the classical results,
starting with \cite{Robbins-Monro51} which introduced the SA algorithm
for the scalar case where $d=1$.
However, we state it here for the multidimensional case.
In that paper, the update equation is \eqref{eq:121}, and
the error $\bxt$ is assumed to satisfy the following assumptions
(though this notation is not used in that paper)
\be\label{eq:211}
E_t ( \bxt ) = \bz , CV_t( \bxt ) \leq M^2
\ee
for some finite constant $M$.
The first assumption implies that $\{ \bxt \}$ is a martingale
difference sequence, and also that
$\f(\bth_t) + \bxt$ is an \textit{unbiased} measurement of $\f(\bth_t)$. 
The second assumption
means that the conditional variance of the error is globally bounded,
both as a function of $\bth_t$ and as a function of $t$.
With the assumptions in \eqref{eq:211}, along with some assumptions
on the function $\f(\cdot)$, it is shown in \cite{Robbins-Monro51}
that $\bth_t$ converges to a solution of $\f(\bths) = \bz$,
provided the step size sequence satisfies the \textbf{Robbins-Monro (RM)}
conditions
\be\label{eq:212}
\sum_{t=0}^\infty \al_t^2 < \infty ,
\sum_{t=0}^\infty \al_t = \infty .
\ee
This approach was extended in \cite{Kief-Wolf-AOMS52}
to finding a stationary point of a $\C^1$
function $J: \R \ap \R$, that is, a solution to
$\gJ(\bth) = \bz$,\footnote{$^4$Strictly speaking,
we should use $J'(\th)$ for the scalar case.
But we use vector notation to facilitate comparison with later formulas.}
using an \textit{approximate gradient} of $J(\cdot)$.
The specific formulation used in \cite{Kief-Wolf-AOMS52} is
\be\label{eq:213}
h_{t+1} := \frac{ J(\th_t + c_t \D + \xi_{t+1}^+) -
J(\th_t - c_t \D + \xi_{t+1}^-)}{2 c_t} 
\approx \gJ(\th_t) .
\ee
where $c_t$ is called the \textbf{increment}, $\D$ is some fixed number,
and $\xi_{t+1}^+$, $\xi_{t+1}^-$ are the measurement errors.
This terminology ``increment'' is not standard but is used here.
As is standard in such a setting, it is assumed that $\gJ(\cdot)$
is globally Lipschitz-continuous with constant $L$.
For simplicity, it is common to assume that these sequences are i.i.d.\
and also independent of each other, with zero mean and finite variance $M^2$.
We too do the same.
In order to make the expression a better and better approximation to
the true $\gJ(\bth_t)$, the increment $c_t$ must approach zero as $\tai$.
Note that there are \textit{two} sources of error in \eqref{eq:213}.
First, even if the errors $\xi^\pm_{t+1}$ are zero, the first-order
difference is not exactly equal to the gradient $\gJ(\bth_t)$.
Second, the presence of the measurement errors $\xi^\pm_{t+1}$
introduces an additional error term.
To analyze this, let us define
\be\label{eq:12ee}
\z_t = E_t(\h_{t+1}) , \x_t = \z_t - \gJ(\bth_t) , \bzt = \h_{t+1} - \z_t .
\ee
In this case, the error term satisfies
\be\label{eq:214}
\nmeu{ E_t ( \bzt) } \leq L c_t ,
CV_t(\bzt) \leq M^2/(2 c_t^2) .
\ee
These conditions are more general than in \eqref{eq:211}.
For this situation, in the scalar case, it was shown in \cite{Kief-Wolf-AOMS52}
that $\bth_t$ converges to a stationary point of $J(\cdot)$
if the Kiefer-Wolfwitz-Blum (KWB) conditions
\be\label{eq:215}
c_t \ap 0 , 
\sum_{t=0}^\infty ( \al_t^2 / c_t^2 ) < \infty ,
\sum_{t=0}^\infty \al_t c_t < \infty ,
\sum_{t=0}^\infty \al_t = \infty 
\ee
are satisfied.
This approach was extended to the multidimensional case in \cite{Blum54},
and it is shown that the same conditions also ensure
convergence when $d > 1$.
Note that the conditions automatically imply the finiteness of
the sum of $\al_t^2$.

Now we summarize subsequent results.
It can be seen from Theorem \ref{thm:22} below that in the present paper,
the error $\bxt$ is assumed to satisfy the following assumptions:
\be\label{eq:216}
\nmeu{E_t(\bxt)} \leq \mu_t (1 + \nmeu{\bth_t} ),
\ee
\be\label{eq:217}
CV_t( \bxt ) \leq M_t^2 (1 + \nmeusq{\bth_t} ) ,
\ee
where $\bth_t$ is the current iteration.
It can be seen that the above assumptions extend \eqref{eq:214} in several ways.
First, the conditional expectation is allowed to grow as an affine function of 
$\nmeu{\bth_t}$, for each fixed $t$.
Second, the conditional variance is also allowed to grow as a quadratic 
function of $\nmeu{\bth_t}$, for each fixed $t$.
Third, while the coefficient $\mu_t$ is required to approach zero,
the coefficient $M_t$ can grow without bound as a function of $t$.
We are not aware of any other paper that makes such general assumptions.
However, there are several papers wherein the assumptions on $\bxt$
are intermediate between \eqref{eq:211} and \eqref{eq:214}.
We attempt to summarize a few of them next.
For the benefit of the reader, we state the results using the notation
of the present paper.

In \cite{Kushner-JMAA77}, the author considers a recursion of the form
\bd
\bth_{t+1} = \bth_t - \al_t \gJ(\bth_t) + \al_t \bxt + \al_t \bbeta_{t+1} ,
\ed
where $\bbeta_t \ap \bz$ as $\tai$.
Here, the sequence $\{ \bxt \}$ is \textbf{not}
assumed to be a martingale difference sequence.
Rather, it is assumed to satisfy a different set of conditions,
referred to as the Kushner-Clark conditions;
see \cite[A5]{Kushner-JMAA77}.
It is then shown that if the error sequence $\{ \bxt \}$ satisfies
\eqref{eq:211}, i.e., is a martingale difference sequence,
then Assumption (A5) holds.
Essentially the same formulation is studied in \cite{Ljung78}.
The same formulation is also studied
\cite[Section 2.2]{Borkar22}, where \eqref{eq:211} holds,
and $\bbeta_t \ap \bz$ as $\tai$.
In \cite{Tadic-Doucet-AAP17}, it is assumed only that
$\limsup_t \bbeta_t < \infty$.
In all cases,
it is shown that $\bth_t$ converges to a solution of $\f(\bths) = \bz$,
\textit{provided} the iterations remain bounded almost surely.
Therefore, the boundedness of the iterations is established via
separate arguments.

In all of the above references, the bound on $CV_t(\bxt)$ is
as in \eqref{eq:211}.
We are aware of only one paper when the bound on $CV_t(\bxt)$
is akin to that in \eqref{eq:217}.
In \cite{IJLL-arxiv23}, the authors study smooth convex optimization.
They assume that the estimated gradient is \textit{unbiased}, so that
$\mu_t = 0$ for all $t$.
However, an analog of \eqref{eq:217} is assumed to hold, which is
referred to as ``state-dependent noise.''
See \cite[Assumption (SN)]{IJLL-arxiv23}.
In short, there is no paper wherein the assumptions on the error are
as general as in \eqref{eq:216} and \eqref{eq:217}.

\subsection{Convergence Theorems}\label{ssec:22}

In this subsection, we state without proof some results from
\cite{MV-RLK-SGD-arxiv23} on the convergence of SA, when
the measurement error satisfies
\eqref{eq:216} and \eqref{eq:217}, which are the most general
assumptions to date.
In addition to proving convergence, we also provide a general framework
for estimating the \textit{rate} of convergence.
The applications of these convergence theorems to stochastic gradient
descent (SGD) are discussed in Section \ref{sec:Opt}.

The theorems proved in \cite{MV-RLK-SGD-arxiv23}
make use of the following classic
``almost supermartingale theorem'' of Robbins-Siegmund
\cite[Theorem 1]{Robb-Sieg71}.
The result is also proved as \cite[Lemma 2, Section 5.2]{BMP92}.
Also see a recent survey paper as \cite[Lemma 4.1]{Fran-Gram22}.
The theorem states the following:

\begin{lemma}\label{lemma:2}
Suppose $\{ z_t \} , \{ f_t \} , \{ g_t \} , \{ h_t \}$ are
stochastic processes taking values in $[0,\infty)$, adapted to some
filtration $\{ \F_t \}$, satisfying
\be\label{eq:221}
E_t( z_{t+1} ) \leq (1 + f_t) z_t + g_t - h_t \as, \fa t ,
\ee
where, as before, $E_t(z_{t+1})$ is a shorthand for $E(z_{t+1} | \F_t )$.
Then, on the set
\bd
\OM_0 := \{ \om : \sum_{t=0}^\infty f_t(\om) < \infty \}
\cap \{ \om : \sum_{t=0}^\infty g_t(\om) < \infty \} ,
\ed
we have that $\lim_{\tai} z_t$ exists, and in addition,
$\sum_{t=0}^\infty h_t(\om) < \infty$.
In particular, if $P(\OM_0) = 1$, then $\{ z_t \}$ is bounded
almost surely, and $\sum_{t=0}^\infty h_t(\om) < \infty$ almost surely.
\end{lemma}

The first convergence result, namely Theorem \ref{thm:21} below,
is a fairly straight-forward, but useful, extension of Lemma \ref{lemma:2}.
It is based on a concept which is introduced in \cite{Gladyshev65}
but without giving it a name.
The formal definition is given in \cite[Definition 1]{MV-MCSS23}:

\begin{definition}\label{def:Class-B}
A function $\eta : \R_+ \ap \R_+$ is
said to \textbf{belong to Class $\B$} if $\eta(0) = 0$, and in addition
\bd
\inf_{\e \leq r \leq M} \eta(r) > 0 , \fa 0 < \e < M < \infty .
\ed
\end{definition}

Note $\eta(\cdot)$ is \textit{not} assumed to be monotonic, or even to be
continuous.
However, if $\eta : \R_+ \ap \R_+$ is continuous, then
$\eta(\cdot)$ belongs to Class $\B$ if and only if (i) $\eta(0) = 0$,
and (ii) $\eta(r) > 0$ for all $r > 0$.
Such a function is called a ``class P function'' in
\cite{Gruene-Kellett14}.
Thus a Class $\B$ function is slightly more general than a function
of Class $P$.

As example of a function of Class $\B$ is given next:
\begin{example}\label{exam:1}
Define a function $f: \R_+ \ap \R$ by
\bd
\phi(\th) = \left\{ \ba{ll} \th, & \mbox{if } \th \in [0,1] , \\
e^{-(\th-1)}, & \mbox{if } \th > 1 . \ea \right.
\ed
Then $\phi$ belongs to Class $\B$.
A sketch of the function $\phi(\cdot)$ is given in Figure \ref{fig:1}.
Note that, if we were to change the definition to:
\bd
\phi(\th) = \left\{ \ba{ll} \th, & \mbox{if } \th \in [0,1] , \\
2 e^{-(\th-1)}, & \mbox{if } \th > 1 , \ea \right.
\ed
then $\phi(\cdot)$ would be discontinuous at $\th = 1$, but it would
still belong to Class $\B$.
Thus a function need not be continuous to belong to Class $\B$.

\bfig
\bc
\includegraphics[width=60mm]{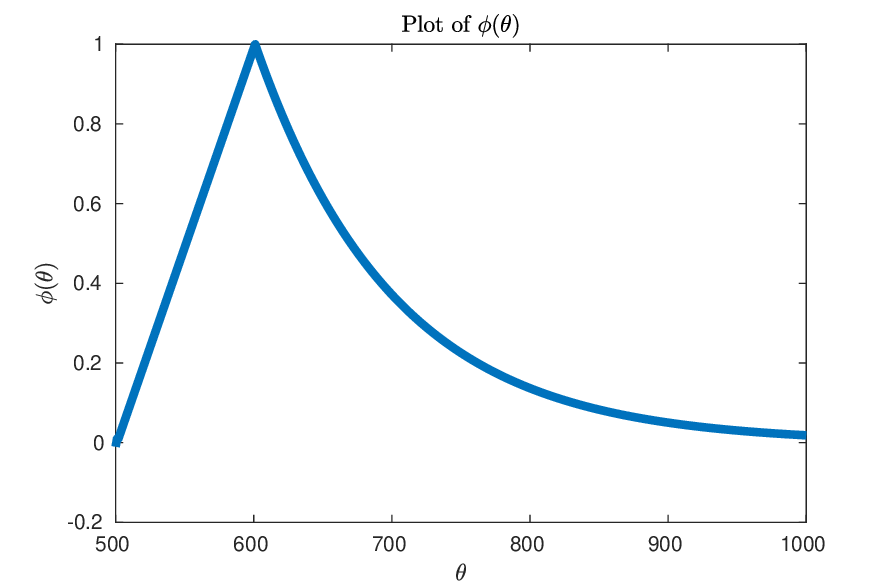}
\ec
\caption{An illustration of a function in Class $\B$}
\label{fig:1}
\efig
\end{example}

Now we present our first convergence theorem, which is
an extension of Lemma \ref{lemma:2}.
This theorem is used to establish the convergence of
stochastic gradient methods for nonconvex functions,
as discussed in Section \ref{sec:Opt}.
It is \cite[Theorem 1]{MV-RLK-SGD-arxiv23}.

\begin{theorem}\label{thm:21}
Suppose $\{ z_t \} , \{ f_t \} , \{ g_t \} , \{ h_t \}, \{ \al_t \}$ are
$[0,\infty)$-valued stochastic processes
defined on some probability space $(\OM,\SI,P)$, and
adapted to some filtration $\{ \F_t \}$.
Suppose further that
\be\label{eq:222}
E_t(z_{t+1} ) \leq (1 + f_t) z_t + g_t - \al_t h_t \as, \fa t .
\ee
Define
\be\label{eq:223}
\OM_0 := \{ \om \in \OM : \sum_{t=0}^\infty f_t(\om) < \infty \mbox{ and }
\sum_{t=0}^\infty g_t(\om) < \infty \} ,
\ee
\be\label{eq:223a}
\OM_1 := \{ \sum_{t=0}^\infty \al_t(\om) = \infty \} .
\ee
Then
\ben
\item
Suppose that $P(\OM_0) = 1$.
Then the sequence $\{ z_t \}$ is bounded almost surely, and
there exists a random variable $W$ defined on $(\OM,\SI,P)$ such that
$z_t(\om) \ap W(\om)$ almost surely.
\item
Suppose that, in addition to $P(\OM_0) = 1$, it is also true that
$P(\OM_1) = 1$.
Then
\be\label{eq:224}
\liminf_{\tai} h_t(\om) = 0  \fa \om \in \OM_0 \cap \OM_1 .
\ee
Further, suppose there exists a function $\eta(\cdot)$ of Class $\B$
such that $h_t(\om) \geq \eta(z_t(\om))$ for all $\om \in \OM_0$.
Then $z_t(\om) \ap 0$ as $\tai$ for all $\om \in \OM_0$.
\een
\end{theorem}

Next we study a linear stochastic recurrence relation.
Despite its simplicity, it is a key tool in establishing the
convergence of Stochastic Gradient Descent (SGD) studied in
Section \ref{sec:Opt}.
Suppose $\bth_0$ is an $\R^d$-valued random variable, and that
$\{ \bzeta_t \}_{t \geq 1}$ is an $\R^d$-valued stochastic process.
Define $\{\bth_t\}_{t \geq 1}$ recursively by
\be\label{eq:225}
\bth_{t+1} = (1 - \alpha_t) \bth_t + \alpha_t \bzt , t \geq 0 ,
\ee
where $\{ \alpha_t \}_{t \geq 0}$ is another $[0,1)$-valued stochastic
process.
Define $\{ \F_t \}$ to be the filtration where
$\F_t$ is the $\s$-algebra generated by $\bth_0, \al_0^t, \bzeta_1^t$.
Note that \eqref{eq:225} is of the form \eqref{eq:122} with
$\gbold(\bth) \equiv \bz$.
Hence $\gbold(\cdot)$ has the unique fixed point $\bz$, and we would
want that $\bth_t \ap \bz$ as $\tai$.
Theorem \ref{thm:22} below is a ready consequence of applying
\cite[Theorem 3]{MV-RLK-SGD-arxiv23} to the function $J(\bth) = 
(1/2)\nmeusq{\bth}$.

\begin{theorem}\label{thm:22}
Suppose there exist sequences of constants $\{ \mu_t \}$,
$\{ M_t \}$ such that, for all $t \geq 0$ we have
\be\label{eq:226}
\nmeu{E_t(\bzt)} = \nmeu{ \boldeta_t} \leq \mu_t (1 + \nmeu{\bth_t} ),
\ee
\be\label{eq:227}
CV_t( \bzt ) = E_t( \nmeusq{\bpsi_{t+1} } ) \leq M_t^2 (1 + \nmeusq{\bth_t} ).
\ee
Under these conditions, if
\be\label{eq:228}
\sum_{t=0}^\infty \alpha_t^2 < \infty, \sum_{t=0}^\infty \mu_t\alpha_t<\infty,
\sum_{t=0}^\infty M^2_t\alpha^2_t<\infty ,
\ee
then $\{ \bth_t \}$ is bounded, and
$\nmeu{\bth_t}$ converges to an $\R^d$-valued
random variable.
If in addition,
\be\label{eq:229}
\sum_{t=0}^\infty \alpha_t = \infty,
\ee
then $\bth_t \ap \bz$.
\end{theorem}

Next, we state an extension of
Theorem \ref{thm:21} that provides an estimate on rates of convergence.
For the purposes of this paper, we use the
following definition inspired by \cite{Liu-Yuan-arxiv22}.

\begin{definition}\label{def:order}
Suppose $\{ Y_t \}$ is a stochastic process, and $\{ f_t \}$
is a sequence of positive numbers.
We say that
\ben
\item $Y_t = O(f_t)$ if $\{ Y_t / f_t \}$ is bounded almost surely.
\item $Y_t = \OM(f_t)$ if $Y_t$ is positive almost surely, and
$\{ f_t/Y_t \}$ is bounded almost surely.
\item $Y_t = \Th(f_t)$ if $Y_t$ is both $O(f_t)$ and $\OM(f_t)$.
\item $Y_t = o(f_t)$ if $Y_t /f_t \ap 0$ almost surely as $\tai$.
\een
\end{definition}

The next theorem is a modification of Theorem \ref{thm:21} that
provides bounds on the \textit{rate} of convergence.
It is \cite[Theorem 2]{MV-RLK-SGD-arxiv23}.

\begin{theorem}\label{thm:21a}
Suppose $\{ z_t \} , \{ f_t \} , \{ g_t \} , \{ \al_t \}$ are
stochastic processes defined on some probability space $(\OM,\SI,P)$,
taking values in $[0,\infty)$, adapted to some
filtration $\{ \F_t \}$.
Suppose further that
\be\label{eq:2210}
E_t(z_{t+1} ) \leq (1 + f_t) z_t + g_t - \al_t z_t \fa t ,
\ee
where
\bd
\sum_{t=0}^\infty f_t(\om) < \infty ,
\sum_{t=0}^\infty g_t(\om) < \infty ,
\sum_{t=0}^\infty \al_t(\om) = \infty .
\ed
Then $z_t = o(t^{-\l})$ for every $\l \in (0,1]$ such that
(i) there exists a $T < \infty$ such that
\be\label{eq:2211}
\al_t(\om) - \l t^{-1} \geq 0 \fa t \geq T ,
\ee
and in addition (ii)
\be\label{eq:2212}
\sum_{t=0}^\infty (t+1)^\l g_t(\om) < \infty ,
\sum_{t=0}^\infty [ \al_t(\om) - \l t^{-1} ] = \infty .
\ee
\end{theorem}

With this motivation, we present a refinement of Theorem \ref{thm:22}.
Again, this is obtained by applying
\cite[Theorem 4]{MV-RLK-SGD-arxiv23} to the function $J(\bth) = (1/2)
\nmeusq{\bth}$.

\begin{theorem}\label{thm:23}
Let various symbols be as in Theorem \ref{thm:22}.
Further, suppose there exist constants $\g > 0$ and $\d \geq 0$ such
that\footnote{$^5$Since $t^{-\g}$ is undefined when $t = 0$,
we really mean $(t+1)^{-\g}$.
The same applies elsewhere also.}
\bd
\mu_t = O(t^{-\g}), M_t = O(t^\d) ,
\ed
where we take $\g = 1$ if $\mu_t = 0$ for all sufficiently large $t$,
and $\d = 0$ if $M_t$ is bounded.
Choose the step-size sequence $\{ \al_t \}$ as
$O(t^{-(1-\phi)})$ and $\OM(t^{-(1-c)})$
where $\phi$ is chosen to satisfy
\bd
0 < \phi < \min \{ 0.5 - \d , \g \} ,
\ed
and $c \in (0,\phi]$.
Define
\be\label{eq:2213}
\nu := \min \{ 1 - 2( \phi + \d) , \g - \phi \} .
\ee
Then $\nmeusq{\bth_t} = o(t^{-\l})$ for every $\l \in (0,\nu)$.
In particular, if $\mu_t = 0$ for all $t$ and $M_t$ is bounded with
respect to $t$, then we can take $\nu = 1 - 2 \phi$.
\end{theorem}

\section{Applications to Stochastic Gradient Descent}\label{sec:Opt}

In this section, we reprise some relevant results from
\cite{MV-RLK-SGD-arxiv23} on the convergence of the Stochastic
Gradient Method.
Specifically, we analyze the convergence of the Stochastic
Gradient Descent (SGD) algorithm in the form
\be\label{eq:311}
\bth_{t+1} = \bth_t - \al_t \h_{t+1} ,
\ee
where $\h_{t+1}$ is a stochastic gradient.
For future use, let us define
\be\label{eq:312}
\z_t = E_t(\h_{t+1}) , \x_t = \z_t - \gJ(\bth_t) , \bzt = \h_{t+1} - \z_t .
\ee
The last equation in \eqref{eq:312} implies that $E_t(\bzt) = \bz$.
Therefore 
\be\label{eq:312a}
E_t( \nmeusq{\h_{t+1}} ) = \nmeusq{\z_t} + E_t \nmeusq{\bzt} .
\ee
We make two assumptions about the stochastic gradient:
\textbf{Assumption:}
There exist sequences of constants $\{ \mu_t \}$ and $\{ M_t \}$ such that
\be\label{eq:313}
\nmeu{\x_t} \leq \mu_t [ 1 + \nmeu{\gJt} ] , \fa \bth_t \in \R^d , \fa t ,
\ee
\be\label{eq:314}
E_t ( \nmeusq{\bzt} \leq M_t^2 [ 1 + J(\bth_t) ] ,
\fa \bth_t \in \R^d , \fa t .
\ee
As mentioned above, these are the least restrictive assumptions in
the literature.

In order to analyze the convergence of \eqref{eq:311},
we make two standing assumptions on $J(\cdot)$, namely:
\ben
\item[(S1)] $J(\cdot)$ is $\C^1$, and $\gJ(\cdot)$ is globally
Lipschitz-continuous with constant $L$.
\item[(S2)] $J(\cdot)$ is bounded below, and the infimum is attained.
Thus
\bd                                      
J^* := \inf_{\bth \in \R^d} J(\bth)
\ed
is well-defined, and $J^* > -\infty$.                  
Moreover, the set
\bd
S_J := \{ \bth : J(\bth = J^* \}
\ed
is nonempty.
Note that hereafter we take $J^* = 0$.
\een

Aside from these standing assumptions, we introduce four other conditions.
Note that not all conditions are assumed in every theorem

\ben
\item[(GG)] There exists a constant $H < \infty$ such that
\bd
\nmeusq{ \gJ(\bth) } \leq H J(\bth) , \fa \bth \in \R^d .
\ed
\item[(PL)] There exists a constant $K$ such that
\bd
\nmeusq{\gJ(\bth)} \geq K J(\bth) , \fa \bth \in \R^d .
\ed
\item[(KL')] There exists a function $\psi(\cdot)$ of Class $\B$
such that
\bd
\nmeu{\gJ(\bth)} \geq \psi(J(\bth) , \fa \bth \in \R^d .
\ed
\item[(NSC)] There exists a function $\eta(\cdot)$ of Class $\B$ such that
\bd
\r(\bth) \leq \eta(J(\bth)) , \fa \bth \in \R^d ,
\ed
where
\bd
\r(\bth) := \inf_{\bphi \in S_J} \nmeu{\bth - \bphi} .
\ed
\een
In the above (GG) stands for ``Gradient Growth.'' It is satisfied
with $H = 2L$ whenever $J(\cdot)$ is convex, but can also hold otherwise.
Condition (PL) stands for ``Polyak-Lojasiewicz,'' while
(KL') stands for ``modified Kurdyka-Lojasiewicz.''
Finally, (NSC) stands for ``Near Strong Convexity.''
A good discussion of (PL) and (KL) (as opposed to (KL'))
can be found in \cite{Karimi-et-al16}, while 
\cite[Section 6]{MV-RLK-SGD-arxiv23} explains the difference between
(KL) and (KL'), as well Condition (NSC).

With this background, we first state a theorem on the convergence of
the SGD, but without any conclusions as to the rate of convergence.
It is \cite[Theorem 3]{MV-RLK-SGD-arxiv23}.

\begin{theorem}\label{thm:31}
Suppose the objective function $J(\cdot)$ satisfies the standing assumptions
(S1) and (S2) together with (GG),
and that the stochastic gradient $\h_{t+1}$ satisfies
\eqref{eq:313} and \eqref{eq:314}.
With these assumptions, we have the following conclusions;
\ben
\item Suppose
\be\label{eq:315}
\sum_{t=0}^\infty \al_t^2 < \infty , 
\sum_{t=0}^\infty \al_t \mu_t  < \infty ,
\sum_{t=0}^\infty \al_t^2 M_t^2 < \infty .
\ee
Then $\{ \gJ(\bth_t) \}$ and $\{ J(\bth_t) \}$ are bounded, and in addition,
$J(\bth_t)$ converges to some random variable as $\tai$.
\item If in addition $J(\cdot)$ satisfies (KL'), and
\be\label{eq:316}
\sum_{t=0}^\infty \al_t = \infty ,
\ee 
then $J(\bth) \ap 0$ and $\gJ(\bth_t) \ap \bz$ as $\tai$.
\item Suppose that in addition to (KL'), $J(\cdot)$ also satisfies (NSC),
and that \eqref{eq:315} and \eqref{eq:316} both hold.
Then $\r(\bth_t) \ap 0$ as $\tai$.
\een
\end{theorem}

Theorem \ref{thm:31} does not say anything about the \textit{rate}
of convergence.
By strengthening the hypothesis from (PL) to (KL'), we can serive
explicit bounds on the rate.
It is \cite[Theorem 4]{MV-RLK-BASA-arxiv21}.

\begin{theorem}\label{thm:32}
Let various symbols be as in Theorem \ref{thm:31}.
Suppose $J(\cdot)$ satisfies the standing assumptions (S1) through (S3),
and also property (PL),
and that \eqref{eq:315} and \eqref{eq:316} hold.
Further, suppose there exist constants $\g > 0$ and $\d \geq 0$ such
that\footnote{$^6$Since $t^{-\g}$ is undefined when $t = 0$,
we really mean $(t+1)^{-\g}$.
The same applies elsewhere also.}
\bd
\mu_t = O(t^{-\g}), M_t = O(t^\d) ,
\ed
where we take $\g = 1$ if $\mu_t = 0$ for all sufficiently large $t$,
and $\d = 0$ if $M_t$ is bounded.
Choose the step-size sequence $\{ \al_t \}$ as
$O(t^{-(1-\phi)})$ and $\OM(t^{-(1-C)})$
where $\phi$ and $C$ are chosen to satisfy
\bd
0 < \phi < \min \{ 0.5 - \d , \g \} , C \in (0,\phi] .
\ed
Define
\be\label{eq:3112}
\nu := \min \{ 1 - 2( \phi + \d) , \g - \phi \} .
\ee
Then $\nmeusq{\gJt} = o(t^{-\l})$ and $J(\bth_t) = o(t^{-\l})$
for every $\l \in (0,\nu)$.
In particular, by choosing $\phi$ very small, it follows that
$\nmeusq{\gJt} = o(t^{-\l})$ and $J(\bth_t) = o(t^{-\l})$ whenever
\be\label{eq:3113}
\l < \min \{ 1 - 2 \d , \g \} .
\ee
\end{theorem}

\begin{corollary}\label{coro:31}
Suppose all hypotheses of Theorem \ref{thm:32} hold.
In particular, if $\mu_t = 0$ for all large enough $t$ in \eqref{eq:313},
and $M_t$ in \eqref{eq:314} is bounded with respect to $t$, then
$\nmeusq{\gJt} = o(t^{-\l})$ and $J(\bth_t) = o(t^{-\l})$ for all $\l < 1$.
\end{corollary}

It is worthwhile to compare the content of Corollary \ref{coro:31}
with the bounds from \cite{Arjevani-et-al-MP23}.
In that paper, it is assumed that $\z_t := E_t(\h_{t+1}) = \gJ(\bth_t)$,
and that $CV_t(\h_{t+1}) \leq M^2$ for some finite constant $M$;
see \cite[Eq.\ (2)]{Arjevani-et-al-MP23}.
In the present notation, this is the same as saying that $\mu_t = 0$ for
all $t$, and that $M_t = M$ for all $t$.
Thus the assumption is that the stochastic gradient $\h_{t+1}$ is unbiased
and has conditional variance that is uniformly bounded with respect to
$t$ and $\bth_t$.
With these assumptions on the stochastic gradient, it is shown 
in \cite{Arjevani-et-al-MP23} that
for an arbitrary convex obective function, the best achievable rate 
$\nmeu{\gJt} = O(t^{-1/2})$, or equivalently, $\nmeusq{\gJt} = O(t^{-1})$.
Thus the bounds in Corollary \ref{coro:31} are tight for any class
of functions satisfying the hypotheses therein, which includes both
convex as well as a class of nonconvex functions.

\section{Block Asynchronous Stochastic Approximation}\label{sec:BASA}

Until now, we have reviewed some results from a companion  paper
\cite{MV-RLK-SGD-arxiv23}.
This section and the next
contain original results due to the authors that are not
reported anywhere else.
Suppose one wishes to solve \eqref{eq:122}, that is, to find a fixed
point of a given map $\gbold(\cdot)$.
As mentioned earlier, when every component of $\bth_t$ is updated at
each $t$, this is the standard version of SA, referred to by us as
``synchronous'' SA, though the term is not very standard.
When \textit{exactly one} component of $\bth_t$ is updated at each $t$,
this is known as ``Asynchronous'' SA, a term first introduced in
\cite{Tsi-ML94}.
In this section, we study the solution of \eqref{eq:122} using
``Block Asynchronous'' SA, whereby, 
At each step $t$, \textit{some but not necessarily all} components of
$\bth_t$ are updated.
This is denoted by the acronym BASA.
Clearly both Synchronous SA and Asynchronous SA are special cases of BASA.

\subsection{Intermittent Updating: Convergence and Rates}\label{ssec:41}

The key distinguishing feature of BASA is that each component of
$\bth_t$ gets updated in an ``intermittent'' fashion.
Before tackling the convergence of BASA in $\R^d$,
in the present subsection we state and prove results analogous to
Theorems \ref{thm:22} and \ref{thm:23} for the scalar case
with intermittent updating.

The problem setup is as follows:
The recurrence relationship is
\be\label{eq:411}
w_{t+1} = (1 - \al_t \kappa_t) w_t + \al_t \kappa_t \xi_{t+1} ,
\ee
where $\{ w_t \}$ is an $\R$-valued stochastic process of interest,
$\{ \xi_t \}$ is the measurement error (or ``noise''),
$\{ \al_t \}$ is a $(0,1)$-valued stochastic process called the
``step size'' process, and $\{ \kappa_t \}$ is a $\bi$-valued
stochastic process called the ``update'' process.
Clearly, if $\kappa_t = 0$, then $w_{t+1} = w_t$, irrespective of
the value of $\al_t$;
therefore $w_{t+1}$ is updated only at those $t$ for which $\kappa_t = 1$.
This is the rationale for the name.
With the update process $\{ \kappa_t \}$, as before we associate a
``counter'' process $\{ \nu_t \}$, defined by
\be\label{eq:412}
\nu_t = \sum_{s = 0}^t \kappa_s .
\ee
Thus $\nu_t$ is the number of times up to and including time $t$
at which $w_t$ is updated.
We also define
\be\label{eq:413}
\nuit := \min \{ t \in \Nbb : \nu_t = \t \} , \fa \t \geq 1 .
\ee
Then $\nu^{-1}(\cdot)$ is well-defined, and
\be\label{eq:414}
\nu(\nuit) = \t , \nu^{-1}(\nu_t) \leq t ,
\nuit \leq \t - 1 .
\ee
The last inequality arises from the fact that there are $t+1$ terms
in \eqref{eq:412}.
Also, $\kappa_t = 1$ only when $t = \nu^{-1}(\t)$ for some $\t$,
and is zero for other values of $t$.
Hence, in \eqref{eq:411}, if $t = \nu^{-1}(\t)$ for some $\t$,
then $w_t$ gets updated to $w_{t+1}$, and
\be\label{eq:414a}
w_{t+1} = w_{t+2} = \cdots = w_{\nu^{-1}(\t+1)},
\ee
at which time $w$ gets updated again.
Thus $w_t$ is a ``piecewise-constant'' process, remaining constant
between updates.
This suggests that we can transform the independent variable from $t$ to $\t$.
Define
\be\label{eq:414c}
x_\t := w_{\nuit} , \zeta_{\t+1} := \xi_{\nuit + 1 } , \fa \t \geq 1 ,
\ee
with the convention that $x_1 = w_0$.
Note that the convention is consistent whether $\nu_0 = 1$ or not
(as can be easily verified).
Also we define
\bd
b_\t := \al_t \kappa_t, 
\ed
whenever $t = \nuit$ for some $\t$.
With these definitions, \eqref{eq:411} is equivalent to
\be\label{eq:415}
x_{\t+1} = ( 1 - b_\t ) x_\t + b_\t \zeta_{\t+1} , \fa \t \geq 1  ,
\ee

Note that, in \eqref{eq:415}, $b_\t$ is a random variable
for all $\t \geq 1$, and that there is no $b_0$.
To analyze the behavior of \eqref{eq:415},
we introduce some preliminary concepts.
Let $\F_t$ be the $\s$-algebra generated by $w_0, \kappa_0^t, \xi_1^t$.
With the change in time indices, define 
$\{ \G_\t \}$, where $\G_\t = \F_{\nuit}$, whenever $t = \nuit$ for some $\t$.
Then it is easy to see that $\{ \G_\t \}$ is also a filtration, and that
\bd
E( x_\t | \G_\t ) = E_t( w_t | \F_t ) 
\ed
whenever $t = \nuit$ for some $\t$.
Hence we can mimic the earlier notation and denote $E ( X | \G_\t)$
by $E_\t(X)$.
Also, if it is assumed that
original step size $\al_t$ belongs to $\M(\F_t)$,
then $b_\t \in \M(\F_t) = \M(\F_{\nuit}) = \M(\G_\t)$.
The assumption implies that, while the step $\al_t$ may be random,
it only makes use of the information available up to and including step $t$.

Now we present a general convergence result for \eqref{eq:415}.
Observe that $\{ w_t \}$ is a ``piecewise-constant version'' of $\{ x_\t \}$.
Hence if some conclusions are established for the $x$-process,
they are also established for the $w$-process, after adjusting for
the time change from $t$ to $\t$.

\begin{theorem}\label{thm:41}
Consider the recursion \eqref{eq:415}.
Suppose there exist constants $\mu_t , M_t$ such that
\be\label{eq:415a}
| E_t(\xi_{t+1} ) | \leq \mu_t (1 + | w_t| ) \fa t \geq 0 ,
\ee
\be\label{eq:415b}
CV_t(\xi_{t+1}) \leq M_t^2 (1 + w_t^2 ) , \fa t \geq 0 .
\ee
Define
\be\label{eq:418}
f_\t = b_\t^2 ( 1 + 2 \mu_{\nuit}^2 + M^2_{\nuit} ) + 3 b_\t \mu_{\nuit} ,
\ee
\be\label{eq:418a}
g_\t = b_\t^2 ( 2 \mu_{\nuit}^2 + M^2_{\nuit} ) + b_\t \mu_{\nuit} .
\ee
Then we have the following conclusions:
\ben
\item If
\be\label{eq:419}
\sum_{\t=1}^\infty f_\t < \infty ,
\sum_{\t=1}^\infty g_\t < \infty ,
\ee
then $x_\t$ is bounded almost surely.
\item If, in addition to \eqref{eq:419}, we also have
\be\label{eq:4110}
\sum_{\t=1}^\infty b_\t = \infty ,
\ee
then $x_\t \ap 0$ as $\t \ap \infty$.
\item If both \eqref{eq:419} and \eqref{eq:4110} hold, then
$x_\t = o(\t^{-\l})$ for every $\l < 1$ such that
\be\label{eq:4111}
\sum_{\t=1}^\infty (\t+1)^{\l} g_\t < \infty ,
\ee
\be\label{eq:4112}
\sum_{\t=1}^\infty [ b_\t - \l \t^{-1} ] = \infty ,
\ee
and in addition, there exists a $T < \infty$ such that
\be\label{eq:4113}
b_\t - \l \t^{-1} \geq 0 \fa \t \geq T .
\ee
\een
\end{theorem}

\begin{proof}
The proof consists of reformulating the bounds on the error
$\bxt$ in such a way that Theorems \ref{thm:22} and \ref{thm:21a} apply.
By assumption, we have that
\bd
| E_t(\xi_{t+1} ) | \leq \mu_t (1 + | w_t| ) \fa t .
\ed
In particular, when $t = \nuit$, we have that $\zeta_{\t+1} = \xi_{t+1}$, and
\bd
| E_\t(\zeta_{\t+1}) | = | E_t(\xi_{t+1} ) | \leq \mu_t (1 + | w_t| ) 
= \mu_{\nuit} ( 1 + | x_\t |) .
\ed
It follows in an entirely analogous manner that
\bd
CV_\t( \zeta_{\t+1} ) \leq M_{\nuit} ( 1 + x_\t^2 ) .
\ed
With these observations, we see that Theorems \ref{thm:22} and \ref{thm:21a}
apply to \eqref{eq:415}, with the only changes being that (i) the
stochastic process is scalar-valued and not vector-valued, (ii)
the time index is denoted by $\t$ and not $t$, and (iii)
$\mu_t, M_t$ are replaced by $\mu_{\nuit}, M_{\nuit}$ respectively.
Now the conclusions of the theorem follow from Theorems
\ref{thm:22} and \ref{thm:21a}.
\end{proof}

Now, for the convenience of the reader, we reprise the two commonly used
approaches for choosing the step size, 
known as a ``global clock'' and a ``local clock'' respectively.
This distinction was apparently first introduced in \cite{Borkar98}.
In each case, there is a \textit{deterministic} sequence
$\{ \beta_t \}_{t \geq 0}$ of step sizes.
If a global clock is used, then $\al_t = \beta_t$ at each update, so that
$b_\t = \beta_{\nuit}$.
If a local clock is used, then $\al_t = \beta_{\nu_t}$, so that
then $b_\t = \beta_{\t-1}$ .
The extra $-1$ in the subscript is to ensure consistency in notation.
To illustrate, suppose $\kappa_t = 1$ for all $t$.
Then $\nu_t = t+1$, and $\nuit = \t-1$.

Now we begin our analysis of \eqref{eq:415} with the two types
of clocks.
Now that Theorem \ref{thm:41} is established,
the challenge is to determine when \eqref{eq:4110} through \eqref{eq:4113}
(as appropriate) hold
for the two choices of step sizes, namely global vs.\ local clocks.

Towards this end, we introduce a few assumptions regarding the update process.
\ben
\item[(U1)] $\nu_t \ap \infty$ as $\tai$ almost surely.
\item[(U2)] There exists a random variable $r$ such that
\be\label{eq:4115}
\frac{\nu_t}{t} \ap r \mbox{ as } \tai , \as .
\ee
\een
Observe that both assumptions are sample-pathwise.
Thus (U2) implies (U1).

We begin by stating the convergence results when a local clock is used.

\begin{theorem}\label{thm:42}
Suppose a local clock is used, so that $\al_t = \beta_{\nu_t}$, so that
$b_\t = \beta_{\t-1}$.
Suppose further that Assumption (U1) holds, and moreover
\ben
\item[(a)]
$\{ \mu_t \}$ is nonincreasing; that is, $\mu_{t+1} \leq \mu_t , \fa t $.
\item[(b)] $M_t$ is uniformly bounded, say by $M$.
\een
With these assumptions,
\ben
\item If
\be\label{eq:4116}
\sum_{t=0}^\infty \beta_t^2 < \infty ,
\sum_{t=0}^\infty \beta_t \mu_t < \infty ,
\ee
then $\{ x_\t \}$ is bounded almost surely, and $\{ w_t \}$ is bounded
almost surely.
\item If, in addition
\be\label{eq:4117}
\sum_{t=0}^\infty \beta_t = \infty ,
\ee
then $x_\t \ap 0$ as $\tai$ almost surely, and $w_t \ap 0$ as $\tai$
almost surely.
\item
Suppose $\beta_t = O(t^{-(1-\phi)})$, for some $\phi > 0$,
and $\beta_t = \OM(t^{-(1-C)})$ for some $C \in (0,\phi]$.
Suppose that $\mu_t = O(t^{-\e})$ for some $\e > 0$.
Then $x_\t \ap 0$ as $\t \ap \infty$, and
$w_t \ap 0$ as $\tai$, for all $\phi < \min \{ 0.5, \e \}$.
Further, $x_\t =  o(\t^{-\l})$, and
$w_t = o((\nu_t)^{-\l})$ for all $\l < \e - \phi$.
In particular, if $\mu_t = 0$ for all $t$, then
$x_\t =  o(\t^{-\l})$, and $w_t = o((\nu_t)^{-\l})$ for all $\l < 1$.
\item If Assumption (U2) holds instead of (U1), then in the previous item,
$w_t = o((\nu_t)^{-\l})$ can be replaced by $w_t = o(t^{-\l})$.
\een
\end{theorem}


\begin{proof}
The proof consists of showing that, under the stated
hypotheses, the appropriate conditions in
\eqref{eq:419} through \eqref{eq:4113} hold.

Recall that $b_\t = \beta_{\t-1}$.
Also, by Assumption (U1), $\nu_t \ap \infty$ as $\tai$, almost surely.
Hence $\nuit$ is well-defined for all $\t \geq 1$.

Henceforth all arguments are along a particular sample path, and
we omit the phrase ``almost surely,'' and also do not display the
argument $\om \in \OM$.

We first prove Item 1 of the theorem.
Recall the definitions of $f_\t$ and $g_\t$ from \eqref{eq:418}
and \eqref{eq:418a} respectively.
Item 1 is established if t is shown that \eqref{eq:419} holds.
For this purpose, note that $\mu_s \leq \mu_t$ if $s > t$, and
$M_t \leq M$ for all $t$.
We analyze each of the three terms comprising $f_\t$.
First,
\bd
\sum_{\t=1}^\infty b_\t^2 = \sum_{\t=1}^\infty \beta_{\t-1}^2
= \sum_{t=0}^\infty \beta_t^2 < \infty .
\ed
Next, since $M_t \leq M$ for all $t$, we have that
\bd
\sum_{\t=1}^\infty b_\t^2 M_{\nuit}^2 \leq
M^2 \sum_{\t=1}^\infty b_\t^2 < \infty .
\ed
Finally,
\bd
\sum_{\t=1}^\infty b_\t \mu_{\nuit}
\leq \sum_{\t=1}^\infty \beta_{\t-1} \mu_{\t-1}
= \sum_{t=0}^\infty \beta_t \mu_t < \infty .
\ed
Here we use the fact that $\nuit \geq \t-1$, so that
$\mu_{\nuit} \leq \mu_{\t-1}$.
Thus it follows from \eqref{eq:418} that $\{ f_\t \} \in \ell_1$,
which is the first half of \eqref{eq:419}.
Next, since $\{ b_\t \mu_{\nuit} \} \in \ell_1$, so is
$\{ b_\t^2 \mu_{\nuit}^2 \} $.
Hence it follows from \eqref{eq:418a} that $\{ g_\t \} \in \ell_1$,
which is the second half of \eqref{eq:419}.
This establishes that $\{ x_\t \}$ is bounded, which in turn implies
that $\{ w_t \}$ is bounded.

To prove Item 2, note that
\bd
\sum_{\t=1}^\infty b_\t = \sum_{\t=0}^\infty \beta_\t = \infty.
\ed
Hence \eqref{eq:4110} holds, and $x_\t \ap 0$ as $\t \ap \infty$,
which in turn implies that $w_t \ap 0$ as $\tai$.

Finally we come to the rates of convergence.
Recall that $\mu_t = O(t^{-\e})$ while $M_t$ is bounded by $M$.
Also, $\beta_t$ is chosen to be $O(t^{-(1-\phi)})$ and $\OM(t^{-(1-C)})$.
From the above, it is clear that
\bd
f_\t = O(\t^{-2 + 2 \phi}) + O(\t^{-1 + \phi - \e }) .
\ed
Hence \eqref{eq:419} holds if
\bd
-2 + 2 \phi < -1 \mbox{ and } -1 + \phi - \e < -1 ,
\mbox{ or } \phi < \min \{ 0.5 , \e \} .
\ed
Next, from the definition of $g_\t$ in \eqref{eq:418a}, it follows that
\bd
(\nuit+1))^\l g_\t \leq (\nu^{-1}(\t+1))^\l g_\t
= O(\t^{-1 + \phi - \e + \l}) .
\ed
Hence \eqref{eq:4111} holds if
\bd
-1 + \phi - \e + \l < -1 \imp \l < \e - \phi .
\ed
Combining everything shows that $x_\t = o(\t^{-\l})$
whenever
\bd
\phi < \min \{ 0.5 , \e \} , \l < \e - \phi .
\ed
If $\mu_t = 0$ for all $t$, then $\e$ can be chosen to be arbitrarily large.
However, the limiting factor is that the argument in Theorem \ref{thm:21a}
holds only for $\l \leq 1$.
Hence $x_\t = o(\t^{-\l})$
whenever
\bd
\phi < 0.5 , \l < 1 .
\ed
Now suppose Assumption (U2) holds, and fix some $\e > 0$.
Then along almost all sample paths, for sufficiently large $T$ we have
that $\nu_t/t \geq r - \e$ for all $t \geq T$.
Thus, whenever $t \geq T$, we have that
\bd
\nu_t \geq rt \imp o((\nu_t)^{-\l}) \leq o((rt)^{-\l}) = o(t^{-\l}) .
\ed
Thus $w_t$ has the same rate of convergence as $x_\t$.
\end{proof}

Since the analysis can commence after a finite number of iterations,
it is easy to see that Assumption (a) above can be replaced by the following:
$\{ \mu_t \}$ is eventually nonincreasing; that is, there
exists a $T < \infty$ such that
\bd
\mu_{t+1} \leq \mu_t , \fa t \geq T .
\ed

Next we state a result when a global clocks is used.
Theorem \ref{thm:43} below is not directly comparable to Theorem
\ref{thm:42} above.
Specifically, in Theorem \ref{thm:42}, the bias coefficient $\mu_t$
is assumed to be non increasing, and the variance bound $M_t^2$
is assumed to bounded uniformly with respect to $t$.
However, the step sizes are constrained only by the requirement
that various summations are finite.
In contrast, in Theorem \ref{thm:43}, there are no assumptions
regarding $\mu_t$ and $\M_t$, but the step size sequence $\{ \beta_t \}$
is assumed to be nonincreasing.

\begin{theorem}\label{thm:43}
Suppose a global clock is used, so that $\al_t = \beta_t$ whenever
$t = \nuit$ for some $\t$ and as a result $b_\t = \beta_{\nuit}$.
Suppose further that Assumption (U2) holds.
Finally, suppose that $\beta_t$ is nonincreasing, so that
$\beta_{t+1} \leq \beta_t$ for all $t$.
$\beta_{t+1} \leq \beta_t$, for all $t$.
Under these assumptions,
\ben
\item If \eqref{eq:4116} holds, and in addition
\be\label{eq:4118}
\sum_{t=0}^\infty \beta_t^2 M_t^2 < \infty ,
\ee
then $\{ w_t \}$ is bounded almost surely.
\item If, in addition, \eqref{eq:4117} holds,
then $w_t \ap 0$ as $\tai$ almost surely.
\item Suppose in addition that
$\beta_t = O(t^{-(1-\phi)})$, for some $\phi > 0$,
and $\beta_t = \OM(t^{-(1-C)})$ for some $C \in (0,\phi]$.
Suppose that $\mu_t = O(t^{-\e})$ for some $\e > 0$,
and $M_t = O(t^\d)$ for some $\d \geq 0$.
Then $w_t \ap 0$ as $\tai$ whenever
\bd
\phi < \min \{ 0.5 - \d , \e \} .
\ed
Moreover, $w_t = o(t^{-\l})$ for all $\l < \e - \phi$.
In particular, if $\mu_t = 0$ for all $t$, then
$w_t = o(t^{-\l})$ for all $\l < 1$.
\een
\end{theorem}

The proof of Theorem \ref{thm:43} makes use of the following auxiliary lemma.

\begin{lemma}\label{lemma:41}
Suppose the update process $\{ \kappa_t \}$ satisfies Assumption (U2).
Suppose $\{ \beta_t \}$ is an $\R_+$-valued sequence of deterministic
constants such that $\beta_{t+1} \leq \beta_t$ for all $t$,
and in addition, \eqref{eq:4117} holds.
Then
\be\label{eq:4120}
\sum_{\t=1}^\infty \beta_{\nuit} = \sum_{t=0}^\infty \beta_t \kappa_t
= \infty .
\ee
\end{lemma}

\begin{proof}
We begin by showing that there exists an integer $M$ such that,
whenever $2^k > M$, we have
\be\label{eq:4121}
\frac{1}{2^k} \left( \sum_{t=2^k+1}^{2^{k+1}} \kappa_t \right)
\geq \frac{r}{2} .
\ee
By assumption, the ratio $\nu_t/t \ap r$ as $\tai$, where $r$ could
depend on the sample path (though the dependence on $\om$ is not displayed).
So we can define $\e = r/2$, and choose an integer $M$ such that
\bd
\left| \frac{1}{T} \sum_{t=0}^{T-1} \kappa_t - r \right|
= \left| \frac{1}{T} \sum_{t=0}^{T-1} (\kappa_t - r) \right|
< \frac{\e}{3} , \fa T \geq M .
\ed
Thus, if $2^k > M$, we have that
\begin{eqnarray*}
\left|
\frac{1}{2^k} \sum_{t=2^k+1}^{2^{k+1}} ( \kappa_t - r ) \right|
& \leq &
\left|
\frac{1}{2^k} \sum_{t=1}^{2^{k+1}} ( \kappa_t - r ) \right|
+ \left|
\frac{1}{2^k} \sum_{t=1}^{2^k} ( \kappa_t - r ) \right| \\
& < & \frac{2}{3} \e + \frac{1}{3} \e = \e = \frac{r}{2} .
\end{eqnarray*}
Next, suppose that $\beta_{t+1} \leq \beta_t$ for all $t$.
(If this holds only for all \textit{sufficiently large} $t$, we
just start all the summations from the time when the above holds.)
\begin{eqnarray*}
\sum_{t=0}^\infty \beta_t \kappa_t
& \geq & \sum_{k=1}^\infty
\left( \sum_{t=2^k+1}^{2^{k+1}} \beta_t \kappa_t \right)
\geq \sum_{k=1}^\infty
\left( \sum_{t=2^k+1}^{2^{k+1}} \beta_{2^{k+1}} \kappa_t \right) \\
& = & \sum_{k=1}^\infty \beta_{2^{k+1}}
\left( \sum_{t=2^k+1}^{2^{k+1}} \kappa_t \right)
\geq \sum_{k=1}^\infty \beta_{2^{k+1}} 2^k \frac{r}{2}
= \frac{r}{4} \sum_{k=1}^\infty \beta_{2^{k+1}} 2^{k+1} \\
& = & \frac{r}{4} \sum_{k=1}^\infty
\sum_{t=2^{k+1}+1}^{2^{k+2}} \beta_{2^{k+1}}
\geq \frac{r}{4} \sum_{k=1}^\infty
\sum_{t=2^{k+1}+1}^{2^{k+2}} \beta_t
= \frac{r}{4} \sum_{k=5}^\infty \beta_t = \infty .
\end{eqnarray*}
This is the desired conclusion.
\end{proof}

\begin{proof}
\textbf{Of Theorem \ref{thm:43}:}
Recall that a global clock is used, so that $b_\t = \beta_{\nuit}$.
Hence
\begin{eqnarray*}
\sum_{\t=1}^\infty f_\t & = &
\sum_{\t=1}^\infty [ \beta_{\nuit}^2 + \beta_{\nuit}^2 M_{\nuit}^2
+ \beta_{\nuit} \mu_{\nuit} ] \\
& = & \sum_{t=0}^\infty[ \beta_t^2 + \beta_t M_t^2 + \beta_t \mu_t ]
< \infty
\end{eqnarray*}
Via entirely similar reasoning, it follows that $\{ g_\t \} \in \ell_1$.
Hence \eqref{eq:419} holds, and Item 1 follows.

To prove Item 2, it is necessary to establish \eqref{eq:4110},
which in this case becomes
\bd
\sum_{\t=1}^\infty \beta_{\nuit} 
= \sum_{\t=0}^\infty b_\t = \infty .
\ed
This is \eqref{eq:4110}.
Hence Item 2 follows.

Finally we come to the rates of convergence.
The only difference is that now $M_t = O(t^\d)$ whereas it was
bounded in Theorem \ref{thm:42}.
To avoid tedious repetition, we indicate only the changed steps.
The only change is that now
\bd
f_\t = O(\t^{-2 + 2 \phi}) + O(\t^{-2+ 2 \phi + 2 \d} )
+ O(\t^{-1 + \phi - \e }) .
\ed
Hence \eqref{eq:419} holds if
\bd
-2 + 2 \phi < -1 , -2 + 2 \phi + 2 \d < -1, \mbox{ and } -1 + \phi - \e < -1 ,
\ed
or
\bd
\phi < \min \{ 0.5 - \d , \e \} .
\ed
Next, from the definition of $g_\t$ in \eqref{eq:418a}, it follows that
\bd
(\nuit+1))^\l g_\t \leq (\nu^{-1}(\t+1))^\l g_\t
= O(\t^{-1 + \phi - \e + \l}) .
\ed
Hence \eqref{eq:4111} holds if
\bd
-1 + \phi - \e + \l < -1 \imp \l < \e - \phi .
\ed
Hence $x_\t = o(\t^{-\l})$ and $w_t = o(t^{-\l})$ whenever
\bd
\phi < \min \{ 0.5 - \d , \e \} , \l < \e - \phi .
\ed
If $\mu_t = 0$ for all $t$, then we can choose $\e$ to be arbitrarily large,
and we are left with
\bd
\phi < 0.5 - \d , \l < 1 .
\ed
\end{proof}

\subsection{Boundedness of Iterations}\label{ssec:42}

Next, we give a precise statement of the class of
fixed point problems to be studied.
In this subsection, it is shown that the iterations are bounded
(almost surely), while in the next subsection, the convergence of
the iterations is established, together with the rate of convergence.
The boundedness of the iterations is established under far more general
conditions than the convergence.
More details are given at the appropriate place.

Let $\Nbb$ denote the set of natural numbers including zero, and let
$\h : \Nbb \times (\R^d)^\Nbb \ap (\R^d)^\Nbb$ denote a 
\textbf{measurement function}.
Thus $\h$ maps $\R^d$-valued sequences into $\R^d$-valued sequences.
The objective is to determine a fixed point of this map when
only noisy measurements of $\h$ are available at each time $t$.
Specifically, define
\be\label{eq:423}
\boldeta_t = \h(t,\bth_0^t) .
\ee
Suppose that, at time $t+1$, the learner has access to a vector
$\boldeta_t + \bxt$, where $\bxt$ denotes the measurement error.
The objective is to determine a sequence $\bpi^* \in (\R^d)^\Nbb$
(if it exists) such that
\bd
\h(\bpi^*) = \bpi^* ,
\ed
using only the noise-corrupted measurements of $\boldeta_t$.

To facilitate this, a few assumptions are made regarding the map $\h$.
First, the map $\h$ is assumed to be
\textbf{nonanticipative}\footnote{$^7$In control and system theory, such a
function is also referred to as ``causal.''} 
and to have \textbf{finite memory}.
The nonanticipativeness of $\h$ means that
\be\label{eq:421}
\bth_0^\infty , \bphi_0^\infty \in (\R^d)^\Nbb ,
\bth_0^t = \bphi_0^t \imp \h(\t,\bth_0^\infty) = \h(\t,\bphi_0^\infty) ,
0 \leq \t \leq t .
\ee
In other words, $\h(t,\bth_0^\infty)$ depends only on $\bth_0^t$.
The finite memory of $\h$ means that there exists a finite constant $\D$
which does not depend on $t$, such that
$\h(t,\bth_0^t)$ further depends only on $\bth_{t-\D+1}^t$.
With slightly sloppy notation, this can be written as
\be\label{eq:423f}
\h(t,\bth_0^t) = \h(t, \bth^t_{t-\D+1}) , \fa t \geq \D ,
\fa \bth_0^\infty \in (\R^d)^\Nbb .
\ee
This formulation incorporates the possibility of
``delayed information'' of the form
\be\label{eq:424}
\eta_{t,i} = g_i( \th_1(t-\D_1(t)) , \cdots , \th_d(t-\D_d(t))) ,
\ee
where $\D_1(t) , \cdots , \D_d(t)$ are delays that could depend on $t$.
The only requirement is that each $\D_j(t) \leq \D$ for some finite $\D$.
This formulation is analogous to
\cite[Eq.\ (2)]{Tsi-ML94} and \cite[Eq.\ (1.4)]{Borkar98},
which is slightly more general in that they require only that
$t - \D_i(t) \ap \infty$ as $\tai$, for each index $i \in [d]$.
In particular, if $\h$ is ``memoryless'' in the sense that, for some
function $\gbold : \R^d \ap \R^d$, we have 
\be\label{eq:4310}
\h(t,\bth_0^t) = \gbold(\bth_t) ,
\ee
then we can take $\D = 1$.
Note that, if $\h$ is of the form \eqref{eq:4310}, then the problem 
at hand becomes one of finding a fixed point in $\R^d$ of the map $\gbold$,
gives noisy measurements of $\gbold$ at eath time step.

To proceed further, it is assumed that the measurement function satisfies
the following assumption:
\ben
\item[(F1)] There exist an integer $\D \geq 1$ and a constant $\g \in (0,1)$
such that
\be\label{eq:423b}
\nmi{ \h(t,\bpsi_{t-\D+1}^t) - \h(t,\bphi_{t-\D+1}^t) } \leq
\g \nmi{ \bpsi_{t-\D+1}^t - \bphi_{t-\D+1}^t } ,
\fa t \geq \D , \fa \bpsi_0^\infty ,\bphi_0^\infty \in (\R^d)^\Nbb .
\ee
This assumption means that the map $\bth^t_{t-\D+1} \mapsto
\h(t, \bth^t_{t-\D+1})$ is a contraction with respect to $\nmi{\cdot}$.
In case $\D = 1$ and $\h$ is of the form \eqref{eq:4310}, Assumption
(F1) says that the map $\gbold$ is a contraction.
\een
Now we discuss a few implications of Assumption (F1).
\ben
\item[(F2)]
By repeatedly applying \eqref{eq:423b} over blocks of width $\D$,
one can conclude that
\be\label{eq:423c}
\nmi{ \h(t,\bpsi_{t-\D+1}^t) - \h(t,\bphi_{t-\D+1}^t) } \leq
\g^{\lfloor t/\D \rfloor}
\nmi{\bpsi_0^{\D-1} - \bphi_0^{\D-1}} ,
\fa \bpsi_0^\infty , \bphi_0^\infty \in (\R^d)^\Nbb .
\ee
Therefore, for every sequence $\bphi_0^\infty$,
the iterations $\h(t,\bphi_0^t)$ converge to a unique fixed point 
$\bpi^*$.
In particular, if we let $(\bpi^*)_0^\infty$ denote the sequence whose
value is $\bpi^*$ for every $t$, then it follows that
\be\label{eq:423d}
\nmi{ \h(t,(\bpi^*)_0^t) - \bpi^* } \leq C_0 \g^{\lfloor t/\D \rfloor} ,
\fa t ,
\ee 
for some constant $C_0$.
\item[(F3)] The following also follows from Assumption (F1):
There exist constants $\r < 1$ and $c_1' > 0$ such that
\be\label{eq:423e}
\nmi{\h(t,\bphi_0^t)} \leq \r  \max\{ c_1' , \nmi{\bphi_0^t}\} , \fa
\bphi\in (\R^d)^\Nbb, t\geq 0 .
\ee
\een

In order to determine $\bpi^*$ in (F2), we use BASA.
Specifically, we choose $\bth_0$ as we wish (either deterministically
or at random).
At time $t$, we update $\bth_t$ to $\bth_{t+1}$ according to
\be\label{eq:422}
\bth_{t+1} = \bth_t + \balpha_t \circ [ \boldeta_t + \bxt ] ,
\ee
where $\balpha_t$ is the \textit{vector of step sizes} belonging to
$[0,1)^d$, $\bxt$ is the measurement noise vector belonging to
$\R^d$, and $\circ$ denotes the Hadamard product.
We are interested in studying two questions:
\ben
\item[(Q1)] Under what conditions is the sequence of iterations
$\{ \bth_t \}$ bounded almost surely?
\item[(Q2)]
Under what conditions does the sequence of iterations $\{ \bth_t \}$
converge to $\bpi^*$ as $\tai$?
\een
Question (Q1) is addressed in this subsection, whereas Question (Q2)
is addressed in the next.

In order to study the above two questions, we make some assumptions
about various entities in \eqref{eq:422}.
Let $\F_t$ denote the $\s$-algebra generated by the random variables
$\bth_0$,  $\bxi_1^t$, and $\al_{0,i}^{t,i}$ for $i \in [d]$.
Then it is clear that $\{ \F_t \}$ is a filtration.
As before, we denote $E(X|\F_t)$ by $E_t(X)$.

The first set of assumptions in on the noise.
\ben
\item[(N1)]
There exists a finite constant $c_1'$ and a sequence of constants
$\{ \mu_t \}$ such that
\be\label{eq:425a}
\nmeu{E_t( \bxt )} \leq c_1' \mu_t (1 + \nmi{\bth_0^t}) ,
\fa t \geq 0 .
\ee
\item[(N2)]
There exists a finite constant $c_2'$ and a sequence of constants
$\{ M_t \}$ such that
\be\label{eq:426}
CV_t( \bxt )
\leq c_2' M_t^2 ( 1 + \nmi{\bth_0^t}^2 ) , \fa t \geq 0 ,
\ee
where, as before,
\bd
CV_t( \bxt ) = 
E_t( \nmeusq{ \bxt - E_t( \bxt ) } )
\ed
\een
Before proceeding further, let us compare the conditions \eqref{eq:425a}
and \eqref{eq:426} with their counterparts \eqref{eq:226} and \eqref{eq:227}
in Theorem \ref{thm:22}.
It can be seen that the above two requirements are more liberal (i.e.,
less restrictive) than in Theorem \ref{thm:22}, because the quantity
$\nmeu{\bth_t}$ is replaced by $\nmi{\bth_0^t}$.
Hence, in \eqref{eq:425a} and \eqref{eq:426}, the bounds are more loose.
However, Theorems \ref{thm:44} and \ref{thm:45} in the next subsection
apply only to \textit{contractive} mappings.
Hence Theorems \ref{thm:44} and \ref{thm:45} complement Theorem \ref{thm:22},
and do not subsume it.

The next set of assumptions is on the step size sequence.
\ben
\item[(S1)]
The random step size sequences $\{ \al_{t,i} \}$ and the sequences
$\{\mu_t\}$, $\{M^2_t\}$ and satisfy (almost surely)
\be\label{eq:427}
\sum_{t=0}^\infty \al_{t,i}^2  < \infty,
\sum_{t=0}^\infty M_t^2\al_{t,i}^2 < \infty,
\sum_{t=0}^\infty \mu_t\al_{t,i}  < \infty,
\fa i \in [d] .
\ee
\item[(S2)] The random step size sequence $\{ \al_{t,i} \}$ satisfies
(almost surely)
\be\label{eq:428}
\sum_{t=0}^\infty \al_{t,i}  = \infty, \as,
\fa i \in [d] .
\ee
\een

With these assumptions in place, we state the main result of this subsection,
namely, the almost sure boundedness of the iterations.
In the next subsection, we state and prove the convergence of the iterations,
under more restrictive assumptions.

\begin{theorem}\label{thm:44}
Suppose that Assumptions (N1) and (N2) about the noise sequence, (S1) and (S2)
about the step size sequence, and (F1) about the function $\h$ hold,
and that $\bth_{t+1}$ is defined via \eqref{eq:422}.
Then $\sup_t\nmi{\bth_t}<\infty$ almost surely.
\end{theorem}

The proof of the theorem is fairly long and involves several preliminary
results and observations.

To aid in proving the results, we introduce a sequence of
``renormalizing constants.''
This is similar to the technique used in \cite{Tsi-ML94}.
For $t \geq 0$, define
\be\label{eq:4241}
\Lambda_t :=  \max\{\nmi{\bth_0^t}, c_1' \},
\ee
where $c_1'$ is defined in \eqref{eq:423}.
With this definition, it follows from \eqref{eq:423e} that
$\boldeta_t=\h(t,\bth_0^t)$ satisfies
\be\label{eq:4242}
\nmi{\boldeta_t} \leq \r \Lambda_t , \fa t .
\ee
Define $\bzt = \L_t^{-1} \bxt$ for all $t \geq 0$.
Now observe that $\L_t^{-1} \leq c_1^{-1}$, and
$\L_t^{-1} \leq (\nmi{ \bth_0^t })^{-1}$.
Hence
\be\label{eq:4245}
\nmi{E_t (  \zeta_{t+1,i} )} \leq c_1' \mu_t (c_1^{-1} + 1)
=: c_2 \mu_t ,
\ee
where $c_2 = c_1' (c_1^{-1} + 1)$.
In particular, the above implies that
\be\label{eq:4245a}
|E_t(  \zeta_{t+1,i} )| \leq c_2 \mu_t , \fa t \geq 0 .
\ee
Similarly
\be\label{eq:4246}
CV_t( \zeta_{t+1,i} ) \leq c_3 M_t^2 , \fa t \geq 0 ,
\ee
for some constant $c_3$.

If we compare \eqref{eq:4245} with \eqref{eq:425a}, and \eqref{eq:4245a} with
\eqref{eq:426}, we see that the bounds for the ``modified'' error
$\bzt$ are simpler than those for $\bxt$.
Specifically, the right side of both \eqref{eq:4245} and \eqref{eq:4245a}
are bounded with respect to $\bth_0^t$ for each $t$, though they may
be unbounded as functions of $t$.
In contrast, the right sides of \eqref{eq:425a} an \eqref{eq:426}
are permitted to be functions of $\nmi{\bth_0^t}$.

Though the next result is quite obvious, we state it separately, because
it is used repeatedly in the sequel.
\begin{lemma}\label{lemma:41a}
For $i\in[d]$ and $0\leq s\leq k<\infty$, define
the doubly-indexed stochastic process
\be\label{eq:4247}
D_i(s,k+1)=\sum_{t=s}^k\Bigl[\prod_{r=t+1}^k(1-\al_{r,i}) \Bigr]\al_{t,i}\zeta_{t+1,i},
\ee
where an empty product is taken as $1$.
Then $\{ D_i(s,k) \}$ satisfies the recursion
\be\label{eq:4247a}
D_i(s,k+1) = (1 - \al_{k,i}) D_i(s,k) + \al_{k,i} \zeta_{k+1,i} ,
D_i(s,s) = 0 .
\ee
In the other direction, \eqref{eq:4247} gives a closed-form
solution for the recursion \eqref{eq:4247a}.
\end{lemma}

Recall that $\Nbb$ denotes the set of non-negative integers $\{0,1,2,\ldots,\}$.
The next lemma is basically the same as \cite[Lemma 2]{Tsi-ML94}.

\begin{lemma}\label{lemma:42}
There exists $\OM_1\subset\OM$ with $P(\OM_1)=1$ and $r_1^*:\OM_1\times (0,1) \ap \Nbb $ such that
\begin{equation}\label{eq:4248}
|D_i(s,k+1)(\om)| \leq \e ,
\fa k \geq s \geq r_1^*(\om,\e) . 
\end{equation}
\end{lemma}

\begin{proof}
Let $\e>0$ be given.
It follows from Lemma \ref{lemma:41a} that $D_i$ satisfies the recursion
\bd
D_i(0,t+1)=(1-\al_{t,i})D_i(0,t)+\al_{t,i}\zeta_{t+1,i}
\ed
with $D_i(0,0)=0$.
Let us fix an index $i\in [d]$, and invoke \eqref{eq:4245a} and \eqref{eq:4246}.
Then it follows from \eqref{eq:4246} that
\bd
CV_t(\zeta_{t+1,i}) \leq c_3 M_t^2,
\ed
and \eqref{eq:4245a} also holds.
Now, if Assumptions (S1) and (S2) also hold, then all the hypotheses needed
to apply Theorem \ref{thm:22} are in place.
Therefore $D_i(0,k+1)$ converges to zero almost surely.
This holds for each $i\in[d]$
Therefore, if we define
\bd
\OM_1 = \{\om\in\OM_1: D_i(0,k+1)(\om) \ap 0 \mbox{ as } \tai
\fa i \in [d] \} ,
\ed
then $P(\OM_1)=1.$
We can see that for $\om\in\OM_1$ we can choose
$r_1^*(\om,\e) \text{ such that }  \fa  k\geq r_1^*(\om,\e), i\in [d]$  we have
\bd
|D_i(0,k+1)(\om)| \leq \textstyle\frac{1}{2}\e.
\ed

To proceed further, we suppress the argument $\om$ in the interests 
of clarity.
Observe from \eqref{eq:4247} that, whenever $s \leq k$ we have
\beq
D_i(s,k+1) & = & \sum_{t=s}^k\Bigl[\prod_{r=t+1}^k(1-\al_{r,i}) \Bigr]
\al_{t,i}\zeta_{t+1,i} \\
& = &  \sum_{t=0}^k\Bigl[\prod_{r=t+1}^k(1-\al_{r,i}) \Bigr]
\al_{t,i}\zeta_{t+1,i} 
-  \sum_{t=0}^{s-1}\Bigl[\prod_{r=t+1}^k(1-\al_{r,i}) \Bigr]
\al_{t,i}\zeta_{t+1,i} \\
& = & D_i(0,k+1) 
- \left[ \prod_{r=s}^k(1-\al_{r,i}) \right]
	\sum_{t=0}^{s-1}\Bigl[\prod_{r=t+1}^{s-1}(1-\al_{r,i}) \Bigr] 
\al_{t,i}\zeta_{t+1,i}  \\
& = & D_i(0,k+1) -
\left[ \prod_{r=s}^k(1-\al_{r,i}) \right] 
D_i(0,s) .
\eeq
Since $1 - \al_{r,i} \in (0,1)$ for all $r,i$, it follows that
the product also belongs to $(0,1)$.
Therefore
\bd
| D_i(s,k+1) | \leq | D_i(0,k+1) | + | D_i(0,s) | \leq
\frac{\e}{2} + \frac{\e}{2}= \e .
\ed
This is the desired conclusion.
\end{proof}

\begin{lemma}\label{lemma:43}
There exists $\OM_2\subset\OM$ with $P(\OM_2)=1$ and
$r_2^*:\OM_1\times \Nbb \times (0,1) \ap \Nbb $ such that
\be\label{eq:4252}
\prod_{s=j}^k(1-\al_{s,i}(\om) )\leq \e , \fa \,k\geq r_2^*(\om,j,\e), i\in [d], \om\in \OM_2.
\ee
\end{lemma}

\begin{proof}
In view of the assumption (S2), if we define
\bd
\OM_2 = \left\{\om\in\OM: \sum_{s=j}^\infty \al_{t,i}(\om) = \infty \fa i\in [d] \right\} ,
\ed
then $P(\OM_2)=1$.
For all $\om\in\OM_2$, we have
\bd
\sum_{s=j}^\infty \al_{t,i}(\om) = \infty.
\ed
Using the elementary inequality $(1-x)\le\exp\{-x\}$ for all $x\in [0,\infty)$, it follows that
\bd
\prod_{s=j}^k(1-\al_{t,i}(\om) )\leq \exp \left\{-\sum_{s=j}^k \al_{t,i}(\om)
\right\} .
\ed
Hence for $\om\in\OM_2$, $\prod_{s=j}^k(1-\al_{t,i}(\om) )$ converges to
zero as $\kai$.
Thus we can choose $r_2^*(\om,j,\e)$ with the required property.
\end{proof}

In the rest of this section, we will fix $\omega\in \OM_1\cap\OM_2$,
the functions $r_1^*$, $r_2^*$ obtained in Lemma \ref{lemma:42} and Lemma
\ref{lemma:43} respectively and prove that if (F1) holds, then
$\nmi{\bth_t(\om)}$ is bounded, which proves Theorem \ref{thm:41}.

Let us rewrite the updating rule \eqref{eq:422} as
\be\label{eq:r212}
\th_{t+1,i} = (1-\al_{t,i} )\th_{t,i} + \al_{t,i} ( \eta_{t,i}  + \Lambda_t\zeta_{t+1,i}) ,
i \in [d],\, t \geq 0,
\ee
By recursively invoking \eqref{eq:r212} for $k\in [0,t]$, we get
\be\label{eq:4253}
\th_{t+1,i}=A_{t+1,i}+B_{t+1,i}+C_{t+1,i}
\ee
where
\be\label{eq:4254}
A_{t+1,i} = \Bigl[\prod_{k=0}^t(1-\al_{k,i} )\Bigr]\th_{0,i},
\ee
\be\label{eq:4255}
B_{t+1,i} = \sum_{k=0}^t\Bigl[\prod_{r=k+1}^t(1-\al_{r,i}) \Bigr]\al_{k,i}\eta_{k,i},
\ee
\be\label{eq:4255b}
C_{t+1,i} = \sum_{k=0}^t\Bigl[\prod_{r=k+1}^t(1-\al_{r,i}) \Bigr]\al_{k,i}\Lambda_k\zeta_{k+1,i}.
\ee

\begin{lemma}\label{lemma:44}
For $i\in [d]$ ,
\be\label{eq:4256}
|C_{t+1,i}|\leq \Lambda_t \sup_{0\leq r\leq t}|D_i(r,t+1)|.
\ee
\end{lemma}

\begin{proof}
We begin by establishing an alternate expression for $C_{k,i}$, namely
\be\label{eq:4256a}
C_{t+1,i} = \L_0 D_i(0,t+1) + \sum_{k=1}^t (\L_k - \L_{k-1}) D_i(k,t+1) ,
\ee
where $D_i(\cdot,\cdot)$ is defined in \eqref{eq:4247}.
For this purpose, observe from Lemma \ref{lemma:41a}
that $C_{t+1,i}$ satisfies
\be\label{eq:4256b}
C_{t+1,i} = \L_t \al_{t,i} \zeta_{t+1,i} + (1 - \al_{t,i}) C_{t,i}
= \L_t D_i(t,t+1) + (1 - \al_{t,i}) C_{t,i} ,
\ee
because $\al_{t,i} \zeta_{t+1,i} = D_i(t,t+1)$ due to \eqref{eq:4247a}
with $s = t$.
The proof of \eqref{eq:4256a} is by induction.
It is evident from \eqref{eq:4255b} that
\bd
C_{1,i} = \L_0 \al_{0,1} \zeta_{1,i} = \L_0 D_i(0,1) .
\ed
Thus \eqref{eq:4256a} holds when $t = 0$.
Now suppose by way of induction that
\be\label{eq:4256c}
C_{t,i} = \L_0 D_i(0,t) + \sum_{k=1}^{t-1} (\L_k - \L_{k-1}) D_i(k,t) .
\ee
Using this assumption, and the recursion \eqref{eq:4256b}, we establish
\eqref{eq:4256a}.

Substituting from \eqref{eq:4256c} into \eqref{eq:4256b} gives
\be\label{eq:4256d}
C_{t+1,i} = \L_t D_i(t,t+1) + \L_0 (1 - \al_{t,i}) D_i(0,t)
+ (1 - \al_{t,i}) \sum_{k=1}^{t-1} (\L_k - \L_{k-1}) D_i(k,t) .
\ee
Now \eqref{eq:4247} implies that
\bd
(1 - \al_{t,i}) D_i(k,t) = D_i(k,t+1) - \al_{t,i} \zeta_{t+1,i}
= D_i(k,t+1) - D_i(t,t+1) .
\ed
Therefore the summation in \eqref{eq:4256d} becomes
\begin{eqnarray*}
\sum_{k=1}^{t-1} (\L_k - \L_{k-1}) (1 - \al_{t,i}) D_i(k,t) & = &
\sum_{k=1}^{t-1} (\L_k - \L_{k-1}) D_i(k,t) \\
& - & D_i(t,t+1) \sum_{k=1}^{t-1} (\L_k - \L_{k-1})
= S_1 + S_2 \mbox{ say}.
\end{eqnarray*}
Then $S_2$ is just a telescoping sum and equals
\bd
S_2 = - \L_{t-1} D_i(t,t+1) + \L_0 D_i(t,t+1) .
\ed
The second term in \eqref{eq:4256d} equals
\bd
\L_0 (1 - \al_{t,i}) D_i(0,t) = \L_0 [ D_i(0,t+1) - \al_{t,i} \zeta_{t+1,i} ]
= \L_0 D_i(0,t+1) - \L_0 D_i(t,t+1) .
\ed
Putting everything together and observing that the term $\L_0 D_i(t,t+1)$
cancels out gives
\bd
C_{t+1,i} = \L_0 D_i(0,t+1) + ( \L_t - \L_{t-1}) D_i(t,t+1)
+ \sum_{k=1}^{t-1} (\L_k - \L_{k-1}) D_i(k,t) .
\ed
This is the same as \eqref{eq:4256d} with $t+1$ replacing $t$.
This completes the induction step and thus \eqref{eq:4256a} holds.
Using the fact that $\L_t \geq \L_{t-1}$,
the desired bound \eqref{eq:4256} follows readily.
\end{proof}

\begin{proof} (Of Theorem \ref{thm:41})
As per the statement of the theorem, we assume that (F1) holds.
We need to prove that
\bd
\sup_{t \geq 0} \L_t  < \infty.
\ed
Define
\bd
\d=\min\{\frac{1-\r}{2\r},\frac{1}{2}\} ,
\ed
and observe that, as a consequence, we have that $\r(1+2\d) \leq 1$.
Choose $ r_1^*=r_1^*(\d)$ as in Lemma \ref{lemma:42} such that
\bd
|D_i(s,k+1)| \leq \d \fa k \geq s \geq r_1^* , \fa i\in [d].
\ed
It is now shown that
\be\label{eq:4259}
 \L_t\leq (1+2\d) \L_{r_1^*} \fa t, \fa i\in[d].
\ee
By the monotonicity of $\{ \L_t \}$, it is already known that
$\L_t \leq \L_{r_1^*}$ for $t \leq r_1^*$.
Hence, once \eqref{eq:4259} is established, it will follow that
\bd
\sup_{0\leq t <\infty}\Lambda_t\leq (1+2\d)\Lambda_{r_1^*}.
\ed

The proof of \eqref{eq:4259} is by induction on $t$.
Accordingly, suppose  \eqref{eq:4259} holds for $t\leq k$.
Using \eqref{eq:4256}, we have
\be\label{eq:4260}
|C_{k+1,i}|\leq \d \L_k\leq \L_{r_1^*}\d(1+2\d).
\ee
It is easy to see from its definition  that
\bd
|A_{k+1,i}|\leq \L_{r_1^*} \Bigl[\prod_{s=0}^k(1-\al_{s,i} )\Bigr]
\ed
Using the induction hypothesis that $\L_t\leq (1+2\d)\L_{r_1^*}$ for $t\leq k$,
we have
\bd
\begin{split}
|B_{k+1,i}|&\leq \sum_{s=0}^k\Bigl[\prod_{r=s+1}^k(1-\al_{r,i}) \Bigr]\al_{s,i}|\eta_{s,i}| \\
&\leq  \sum_{s=0}^k\Bigl[\prod_{r=s+1}^k(1-\al_{r,i}) \Bigr]\al_{s,i} \rho\L_s\\
&\leq \rho(1+2\d)\L_{r_1^*} \sum_{s=0}^k\Bigl[\prod_{r=s+1}^k(1-\al_{r,i}) \Bigr]\al_{s,i} \\
&\leq \L_{r_1^*} \sum_{s=0}^k\Bigl[\prod_{r=s+1}^k(1-\al_{r,i}) \Bigr]\al_{s,i} ,
\end{split}
\ed
because $\r(1 + 2\d) \leq 1$.
Also, the following identity is easy to prove by induction.
\be\label{eq:4264}
\Bigl[\prod_{s=0}^k(1-\al_{s,i} )\Bigr]+ \sum_{s=0}^k
\Bigl[\prod_{r=s+1}^k(1-\al_{r,i})\Bigr]\al_{s,i}=1 \fa k <\infty
\ee
Combining these bounds gives
\bd
|A_{k+1,i}|+|B_{k+1,i}|\leq \L_{r_1^*} .
\ed
Combining this with \eqref{eq:4253} and \eqref{eq:4260} leads to
\bd
\th_{k+1,i}\leq \L_{r_1^*}(1+\d(1+2\d))\leq \L_{r_1^*}(1+2\d).
\ed
Therefore $\nmi{\bth_{k+1}} \leq \L_{r_1^*}(1+2\d)$, and
\bd
\L_{k+1} = \max \{ \nmi{\bth_{k+1}} , \L_k \} \leq \L_{r_1^*}(1+2\d) .
\ed
This proves the induction hypothesis
and completes the proof of Theorem \ref{thm:41}.
\end{proof}

\subsection{Convergence of Iterations with Rates}\label{ssec:43}

In this subsection, we further study the iteration sequence \eqref{eq:422},
under a variety of Block (or Batch) updating schemes, corresponding to
various choices of the step sizes.
Whereas the almost sure boundedness of the iterations is established
in the previous subsection, in this subsection we prove that the
iterations converge to the desired fixed point $\bpi^*$.
Then we also find bounds on the rate of convergence.

We study three specific methods for choosing the step size vector
$\balpha_t$ in \eqref{eq:422}.
Within the first two methods, we further divide into local clocks and
global clocks.
However, in the third method, we permit only the use of a global clock,
for reasons to be specified.

\subsubsection{Convergence Theorem}

The overall plan is to follow up Theorem \ref{thm:44}, which establishes
the almost sure boundedness of the iterations, with a stronger result
showing that the iterations converge almost surely to $\bpi^*$, the
fixed point of the map $\h$.
This convergence is established under the same assumptions as in
Theorem \ref{thm:44}.
In particular, the step size sequence is assumed to satisfy (S1) and (S2).
Having done this, we then study conditions under which
(S1) and (S2) hold for each of the three methods for choosing the step sizes.

\begin{theorem}\label{thm:45}
Suppose that Assumptions (N1) and (N2) about the noise sequence, (S1) and (S2)
about the step size sequence, and (F1) about the function $\h$ hold,
and that $\bth_{t+1}$ is defined via \eqref{eq:422}.
Then $\bth_t \ap \bpi^*$ as $\tai$ almost surely, where $\bpi^*$
is defined in (F2).
\end{theorem}

\begin{proof}
From \eqref{eq:4253}, we have an expression for $\bth_{t+1,i}$,
where $A_{t+1,i}$, $B_{t+1,i}$ and $C_{t+1,i}$ are given by
\eqref{eq:4254}, \eqref{eq:4255} and \eqref{eq:4255b} respectively.
Also, by changing notation from $k$ to $t$ and $s$ to $k$
in \eqref{eq:4264}, and multiplying both sides by $\pi^*_i$, we can write
\bd
\pi^*_i = \Bigl[\prod_{k=0}^t(1-\al_{k,i} )\Bigr] \pi^*_i + 
\left\{ \sum_{k=0}^t
\Bigl[\prod_{r=k+1}^t(1-\al_{r,i})\Bigr] \al_{k,i} \right\} \pi^*_i , \fa t .
\ed
Substituting from these formulas gives
\be\label{eq:431}
\th_{t+1,i} - \pi^*_i = \bar{A}_{t+1,i}
+ \bar{B}_{t+1,i} + C_{t+1,i} ,
\ee
where
\be\label{eq:432}
\bar{A}_{t+1,i} = \prod_{k=0}^t(1-\al_{k,i} ) ( \th_{0,i} - \pi^*_i ) ,
\ee
\be\label{eq:433}
\bar{B}_{t+1,i} = \Bigl[\prod_{r=k+1}^t(1-\al_{r,i})\Bigr]\al_{k,i}
( \eta_{k,i} - \pi^*_i ) ,
\ee
and $C_{t+1,i}$ is as in \eqref{eq:4255b}.
It is shown in turn that each of these quantities approaches zero as $\tai$.

First, from Assumption (S2), it follows that\footnote{$^8$We omit the phrase
``almost surely'' in these arguments.}
\bd
\prod_{k=0}^t(1-\al_{k,i} ) \ap 0 \mbox{ as } \tai .
\ed
Since $\th_{0,i} - \pi^*_i$ is a constant along each sample path,
$\bar{A}_{t+1,i}$ approaches zero.

Second, by combining \eqref{eq:423c} and \eqref{eq:423d} in Property (F2),
it follows that
\bd
| \eta_{t,i} - \pi^*_i | \leq \g^{\lfloor t/\D \rfloor}
\nmi{\bth_0^\D - (\bpi^*)^\D }
\leq C_1 \g^{\lfloor t/\D \rfloor} 
\ed
for some constant $C_1$ (which depends on the sample path).
Thus
\bd
\sum_{r=0}^\infty | \eta_{t,i} - \pi^*_i |  < \infty
\ed
along almost all sample paths.
Now it follows from \eqref{eq:433} that
\beq
| \bar{B}_{t+1,i} | & \leq & \Bigl[\prod_{r=k+1}^t(1-\al_{r,i})\Bigr]\al_{k,i}
| \eta_{k,i} - \pi^*_i | \nonumber \\
& \leq & \Bigl[\prod_{r=k+1}^t(1-\al_{r,i})\Bigr]\al_{k,i} 
C_1 \g^{\lfloor t/\D \rfloor} =: L_{t+1,i} .
\eeq
Let $L_{t+1,i}$ denote the right side of this inequality.
Then it follows from Lemma \ref{lemma:41a} that $L_{t+1,i}$ satisfies
the recursion
\be\label{eq:434}
L_{t+1,i} = (1 - \al_{t,i}) L_{t,i} + \al_{t,i}
C_1 \g^{\lfloor t/\D \rfloor} .
\ee
The convergence of $L_{t+1,i}$ to zero can be proved using
Theorem \ref{thm:41}.
Since the quantity $C_1 \g^{\lfloor t/\D \rfloor} $ is deterministic,
its mean is itself and its variance is zero.
So in \eqref{eq:415a} and \eqref{eq:415b}, we can define
\bd
\mu_t^L := C_1 \g^{\lfloor t/\D \rfloor} , M_t^L := 0 \fa t .
\ed
We can substitute these definitions into \eqref{eq:418} and \eqref{eq:418a},
and define
\be\label{eq:434a}
f_\t^L = b_\t^2 ( 1 + 2 \mu_{\nuit}^2  ) + 3 b_\t \mu_{\nuit} ,
\ee
\be\label{eq:434b}
g_\t^L = b_\t^2 ( 2 \mu_{\nuit}^2 ) + b_\t \mu_{\nuit} .
\ee
Since $\al_t \in [0,1]$ and the sequence $\{ \mu_t^L \}$ is summable
(because $\g < 1$), and $M_t^L \equiv 0$, \eqref{eq:419} is satisfied.
Also, by Assumption (S2), \eqref{eq:4110} is satisfied.
Hence $L_{t+1,i} \ap 0$ as $\tai$, which in turn implies that
$\bar{B}_{t+1,i} \ap 0$ as $\tai$.

Finally, we come to $C_{t+1,i}$.
It is evident from \eqref{eq:4255b} and Lemma \ref{lemma:41a} that
$C_{t+1,i}$ satisfies the recursion
\be\label{eq:435}
C_{t+1,i} = (1 - \al_{t,i}) C_{t,i} + \al_{t,i} \L_t \zeta_{t,i} .
\ee
Now observe that $\L_t$ is bounded, and the rescaled error
signal $\zeta_{t+1,i}$ satisfies \eqref{eq:4245a} and \eqref{eq:4246}.
Hence, if $\L^*$ is a bound for $\L_t$, then it follows from
\eqref{eq:4245a} and \eqref{eq:4246} that
\be\label{eq:435a}
|E_t( \L_t \zeta_{t+1,i} )| \leq c_2 \L^*  \mu_t , \fa t \geq 0 ,
CV_t( \L_t \zeta_{t+1,i} ) \leq c_3  \L^* M_t^2 , \fa t \geq 0 ,
\ee
Hence, when Assumptions (S1) and (S2) hold, it follows from
Theorem \ref{thm:41} that $C_{t+1,i} \ap 0$ as $\tai$.
\end{proof}

\subsubsection{Various Types of Updating and Rates of Convergence}

Next, we describe three different ways of choosing the update processes
$\{ \kappa_{t,i} \}$.

\textbf{Bernoulli Updating:}
For each $i \in [d]$, choose a rate $b_i \in (0,1]$, and let
$\{ \kappa_{t,i} \}$ be a Bernoulli process such that 
\bd
\Pr \{ \kappa_{t,i} = 1 \} = b_i , \fa t .
\ed
Moreover, the processes $\{ \kappa_{t,i} \}$ and $\{ \kappa_{t,j} \}$
are independent whenever $i \neq j$.
Let $\nu_{t,i}$, the counter process for coordinate $i$, be defined as usual.
Then it is easy to see that $\nu_{t,i} / t \ap b_i$ as $\tai$,
for each $i \in [d]$.
Thus Assumption (U2) is satisfied for each $i \in [d]$.

\textbf{Markovian Updating:}
Suppose $\{ Y_t \}$ is a sample path of an irreducible Markov process
on the state space $[d]$.
Define the update process $\{ \kappa_{t,i} \}$ by
\bd
\kappa_{t,i} = I_{\{ Y_t = i \} }
= \left\{ \ba{ll} 1, & \mbox{if } Y_t = i , \\
0, & \mbox{if } Y_t \neq i . \ea \right.
\ed
Let $\bmu$ denote the stationary distribution of the Markov process.
Then the ratio $\nu_{t,i} / t \ap \mu_i$ as $\tai$,
for each $i \in [d]$.
Hence once again Assumption (U2)  holds.

\textbf{Batch Markovian Updating:}
This is an extension of the above.
Instead of a single Markovian sample path, there are $N$ different sample
paths, denoted by $\{ Y_t^n \}$ where $n \in [N]$.
Each sample path $\{ Y_t^n \}$ comes an irreducible Markov process over
the state space $[d]$, and the dynamics of different Markov processes
could be different (though there does not seem to be any advantage to doing
this).
The update process is now given by
\bd
\kappa_{t,i} = \sum_{n \in [N]} I_{\{ Y_t^n = i \} } .
\ed
Define the counter process $\nu_{t,i}$ as before,
and let $\bmu^n$ denote the stationary distribution of the $n$-th Markov
process.
Then
\bd
\frac{\nu_{t,i}}{t} \ap \sum_{n \in [N]} \mu_i^n .
\ed
Hence once again Assumption (U2) holds.

Now we establish convergence rates under each of the above updating
methods (and indeed, any method such that Assumption (U2) is satisfied).
The proof of Theorem \ref{thm:45} gives us a hint on how this can be done.
Specifically, each of the entities $\bar{A}_{t+1,i}, L_{t+1,i}, C_{t+1,i}$
satisfies a stochastic recursion, whose rate of convergence can be established
using Theorems \ref{thm:42} and \ref{thm:43}.
These theorems apply to \textit{scalar-valued} stochastic processes
with intermittent updating.
In principle, when updating $\bth_t$, we could use a mixture of global
and local clocks for different components.
However, in our view, this would be quite unnatural.
Instead, it is assumed that for \textit{every} component, either a global
clock or a local clock is used.
Recall also the bounds \eqref{eq:425a} and \eqref{eq:426} on the error
$\bxt$.

\begin{theorem}\label{thm:46}
Suppose a local clock is used, so that $\al_{t,i} = \beta_{\nu_{t,i}}$
for each $i$ that is updated at time $t$.
Suppose that
$\{ \mu_t \}$ is nonincreasing; that is, $\mu_{t+1} \leq \mu_t , \fa t $,
and $M_t$ is uniformly bounded, say by $M$.
Suppose in addition that $\beta_t = O(t^{-(1-\phi)})$, for some $\phi > 0$,
and $\beta_t = \OM(t^{-(1-C)})$ for some $C \in (0,\phi]$.
Suppose that $\mu_t = O(t^{-\e})$ for some $\e > 0$.
Then $\bth_\t \ap 0$ as $\t \ap \infty$ for all $\phi < \min \{ 0.5, \e \}$.
Further, $\bth_\t =  o(\t^{-\l})$ for all $\l < \e - \phi$.
In particular, if $\mu_t = 0$ for all $t$, then
$\bth_\t =  o(\t^{-\l})$ for all $\l < 1$.
\end{theorem}

The proof of the rate of convergence uses Item (3) of Theorem \ref{thm:41}.
In the proof, let us ignore the index $i$ wherever possible, because
the subsequent analysis applies to each index $i$.
Recall that $\bar{A}_{t+1,i}$ is defined in \eqref{eq:432}.
Since $\ln(1 - x) \leq - x$ for all $x \in (0,1)$, it follows that
\bd
\ln \prod_{k=0}^t(1-\al_{k,i} ) \leq - \sum_{k=0}^t \al_{k,i} ,
\ed
where $\al_{k,i} = 0$ unless there is an update at time $k$.
Now, since a local clock is used, we have that $\al_{k,i} = \beta_{\nu_{k,i}}$
whenever there is an update at time $k$.
Therefore
\bd
\sum_{k=0}^t \al_{k,i} = \sum_{s = 0}^{\nu_{t,i}} \beta_s 
\ed
Now, if Assumption (U2) holds (which it does for each of the three
types of updating considered), it follows that $\nu_{t,i} \approx t/r$
for large $t$.
Thus, if $\beta_\t = \OM(\t^{-(1-C)})$, then we can reason as follows:
\bd
\sum_{s = 0}^{\nu_t} \beta_s \approx \sum_{s=0}^{t/r} s^{-(1-C)}
\approx (t/r)^C .
\ed
Therefore, for large enough $t$, we have that
\bd
\prod_{k=0}^t(1-\al_k ) \leq \exp(-(t/r)^C) .
\ed
It follows from \eqref{eq:432} that $\bar{A}_{t+1,i} \ap 0$
\textit{geometrically fast}.

Next we come to $\bar{B}_{t+1,i}$, which is bounded by $L_{t+1,i}$,
as defined in \eqref{eq:434}.
Recall the definitions \eqref{eq:434a} and \eqref{eq:434b}
for the sequences $\{ f_\t^L \}$ and $\{ g_\t^L \}$.
Then \eqref{eq:419} and \eqref{eq:4110} will hold whenever $C > 0$.
Since Assumption (U2) holds, we have that 
\bd
\mu_{\nuit}^L = C_1 \g^{ \lfloor \nuit /\D \rfloor } \leq C_2 \g^{r' \t}
\ed
for suitable constants $C_2$ and $r'$.
The point to note is that the sequence $\{ C_2 \g^{r' \t} \}$ is a
\textit{geometrically convergent} sequence because $\g < 1$.
Therefore \eqref{eq:4111} holds for \textit{every} $\l > 0$.
Also, \eqref{eq:4112} holds for all $C > 0$.
Hence it follows from Item (3) of Theorem \ref{thm:41} that
$L_{t+1,i} = o(t^{-\l})$ for every $\l > 0$.

This leaves only $C_{t+1,i}$.
We already know that $C_{t+1,i}$ satisfies the recursion \eqref{eq:435}.
Moreover, the modified error sequence $\{ \L_t \zeta_{t,i} \}$
satisfies \eqref{eq:435a}.
The estimates for the rate of convergence now follow from Item (3) of
Theorem \ref{thm:41}, and need not be discussed again.

\begin{theorem}\label{thm:47}
Suppose a global clock is used, so that $\al_{t,i} = \beta_{t,i}$ whenever
the $i$-th component of $\bth_t$ is updated.
Suppose that $\beta_t$ is nonincreasing, so that
$\beta_{t+1} \leq \beta_t$ for all $t$.
Suppose in addition that
$\beta_t = O(t^{-(1-\phi)})$, for some $\phi > 0$,
and $\beta_t = \OM(t^{-(1-C)})$ for some $C \in (0,\phi]$.
Suppose that $\mu_t = O(t^{-\e})$ for some $\e > 0$,
and $M_t = O(t^\d)$ for some $\d \geq 0$.
Then $\bth_t \ap 0$ as $\tai$ whenever
\bd
\phi < \min \{ 0.5 - \d , \e \} .
\ed
Moreover, $\bth_t = o(t^{-\l})$ for all $\l < \e - \phi$.
In particular, if $\mu_t = 0$ for all $t$, then
$\bth_t = o(t^{-\l})$ for all $\l < 1$.
\end{theorem}

The proof is omitted as it is very similar to that of Theorem \ref{thm:46}.

\section{Conclusions and Problems for Future Research}\label{sec:Conc}

In this paper, we have reviewed some results on the convergence of the
Stochastic Gradient method from \cite{MV-RLK-SGD-arxiv23}.
Then we analyzed the convergence of ``intermittently updated'' processes
of the form \eqref{eq:411}.
For this formulation, we derived sufficient conditions for convergence,
as well as bounds on the \textit{rate} of convergence.
Building on this, we derived both sufficient conditions for
convergence, and bounds on the \textit{rate} of convergence,
for the full BASA formulation of \eqref{eq:122}.
Next, we applied these results to derive sufficient conditions for the
convergence of a fixed point iteration with noisy measurements.

There are several interesting problems thrown up by the analysis here.
To our knowledge, our paper is the first to provide explicit estimates
of the rates of convergence for BASA.
A related issue is that of ``Markovian'' stochastic approximation,
in which the update process is the sample path of an irreducible Markov
process.
It would be worthwhile to examine whether the present approach can
handle Markovian SA as well.

\section*{Acknowledgements}

The research of MV was supported by the Science and Engineering Research
Board, India.
The authors thank the anonymous reviewer for a very thorough reading of the
previous version and for many helpful comments.

\end{document}